\pdfoutput=1

\documentclass[11pt]{article}
\usepackage[]{acl}

\usepackage{times}
\usepackage{latexsym}
\usepackage{graphicx}
\usepackage{booktabs}
\usepackage{multirow}

\usepackage[T1]{fontenc}

\usepackage[utf8]{inputenc}

\usepackage{microtype}

\title{Multilingual Previously Fact-Checked Claim Retrieval}

\author{Matúš Pikuliak, Ivan Srba, Robert Moro, Timo Hromadka, Timotej Smolen \\\textbf{Martin Melisek, Ivan Vykopal, Jakub Simko, Juraj Podrouzek, Maria Bielikova} \\
  Kempelen Institute of Intelligent Technologies \\
  \texttt{\{name.surname\}@kinit.sk} 
}

\begin{document}
\maketitle

\newcommand{\factChecksTotal}{205,751}
\newcommand{\factChecksRounded}{206k}
\newcommand{\factCheckingOrganizationsTotal}{142}
\newcommand{\factChecksLanguages}{39}
\newcommand{\factChecksAfp}{21,837}
\newcommand{\factChecksAfpCountries}{23}

\newcommand{\socialMediaPostsTotal}{28,092}
\newcommand{\socialMediaPostsRounded}{28k}
\newcommand{\socialMediaPostsLanguages}{27}

\newcommand{\confirmedPairsTotal}{31,305}
\newcommand{\confirmedPairsRounded}{31k}
\newcommand{\confirmedPairsInSameLanguage}{26,774}
\newcommand{\confirmedPairsInDifferentLanguage}{4,212}

\newcommand{\languagesTotal}{27/39}

\begin{abstract}
Fact-checkers are often hampered by the sheer amount of online content that needs to be fact-checked. NLP can help them by retrieving already existing fact-checks relevant to the content being investigated. This paper introduces a new multilingual dataset -- \textit{MultiClaim} -- for previously fact-checked claim retrieval. We collected {\socialMediaPostsRounded} posts in {\socialMediaPostsLanguages} languages from social media, {\factChecksRounded} fact-checks in {\factChecksLanguages} languages written by professional fact-checkers, as well as {\confirmedPairsRounded} connections between these two groups. This is the most extensive and the most linguistically diverse dataset of this kind to date. We evaluated how different unsupervised methods fare on this dataset and its various dimensions. We show that evaluating such a diverse dataset has its complexities and proper care needs to be taken before interpreting the results. We also evaluated a supervised fine-tuning approach, improving upon the unsupervised method significantly.
\end{abstract}

\section{Introduction}
\label{sec:intro}

Fact-checking organizations have made progress in recent years in manually and professionally fact-checking viral content~\citep{micallef_true_2022, full_fact_challenges_2020}. To reduce some of the fact-checkers' manual efforts and make their work more effective, several studies have recently examined their needs and identified tasks that could be automated~\citep{nakov_automated_2021, full_fact_challenges_2020, micallef_true_2022, dierickx_report_2022, hrckova_2022_automated}. These tasks include searching for the source of evidence for verification, searching for other versions of misinformation, and searching within existing fact-checks. They were identified as particularly challenging for fact-checkers working in low-resource languages~\citep{hrckova_2022_automated}.

In this work, we focus on \textit{previously fact-checked claim retrieval} (PFCR)~\cite{shaar-etal-2020-known}. Given a text making an \textit{input claim} (e.g., a social media post) and a set of \textit{fact-checked claims}, our task is to rank the \textit{fact-checked claims} so that those that are the most relevant w.r.t. the \textit{input claim} (and thus the most useful from the fact-checker's perspective) are ranked as high as possible. 

Previously, this task was mostly done in English. Other languages that have been considered include Arabic~\cite{checkthat_2022}, Bengali, Hindi, Malayalam, and Tamil~\cite{kazemi_claim_2021}. However, many other languages or even entire major language families have not been considered at all. Additionally, so far only \textit{monolingual PFCR} has been tackled, when the input claim and the fact-checked claims are in the same language. To address these shortcomings, we introduce in this paper a new extensive multilingual dataset. Our two main contributions are:

\paragraph{1. \textit{MultiClaim} -- Multilingual dataset for PFCR.} We collected and made available\footnote{The dataset is available at Zenodo upon request \textit{for research purposes only}: \url{https://zenodo.org/record/7737983}. The source code is available at: \url{https://github.com/kinit-sk/multiclaim}.} a novel multilingual dataset for PFCR. The dataset consists of {\factChecksTotal} fact-checks in {\factChecksLanguages} languages and {\socialMediaPostsTotal} social media posts (from now on just \textit{posts}) in {\socialMediaPostsLanguages} languages. For most of these languages, this is the first time this task has been considered at all. This is also the biggest dataset of fact-checks released to date.

All the posts were previously reviewed by professional fact-checkers who also assigned appropriate fact-checks to them. We collected these assignments and gathered {\confirmedPairsTotal} pairs consisting of a post and a fact-check reviewing the claim made in the post. {\confirmedPairsInDifferentLanguage} of these pairs are crosslingual (i.e., the language of the fact-check and the language of the post are different). This dataset introduces \textit{crosslingual PFCR} as a new task that has not been tackled before. This is the biggest collection of such pairs that were confirmed by professional fact-checkers. The dataset also includes OCR transcripts of the images attached to the posts and machine translation of all the data into English.
    
\paragraph{2. In-depth multilingual evaluation.} We evaluated the performance of various text embedding models and BM25 for both the original multilingual data and their English translations. We describe several pitfalls related to the complexity of evaluating such a linguistically diverse dataset. We also explore the performance across several other data dimensions, such as post length or publication date. Finally, we show that we can improve text embedding methods further by using supervised training with our data.

\section{Related Work}
\label{sec:relworks}


Other names are used for PFCR or similar tasks for various reasons, e.g., fact-checking URL recommendation~\citep{voRiseGuardiansFactchecking2018}, fact-checked claims detection~\citep{shaar-etal-2020-known}, verified claim retrieval~\citep{checkthat_2020}, searching for fact-checked information~\citep{vo_lee_facts}, or claim matching~\citep{kazemi_claim_2021}.

\paragraph{Datasets.}
\textit{CheckThat!} datasets~\cite{checkthat_2020, checkthat_2021} have the most similar collection approach to ours. They collect English and Arabic tweets mentioned in fact-checks to create preliminary pairs and then manually filter them. Compared to this work, we broaden the scope of data collection and omit the manual cleaning in favor of using fact-checkers' reports. \citet{shaar-etal-2020-known} collected data from fact-checking of English political debates done by fact-checkers. The \textit{CrowdChecked} dataset \cite{crowdchecked_2022} was created by searching for fact-check URLs on Twitter and collecting English tweets from retrieved threads. The process is inherently noisy and, the authors propose different noise filtering techniques.

\citet{kazemi_claim_2021} collected several million chat messages from public chat groups and tiplines in English, Bengali, Hindi, Malayalam, and Tamil and 150k fact-checks. Then they sampled roughly 2,300 pairs based on their embedding similarity and manually annotated them. In the end, they obtained only roughly 250 positive pairs. \citet{jiang_categorising_2021} matched COVID-19 tweets and 90 COVID-19 claims in a similar manner. Their data could be used for PFCR, but the authors worked on classification instead.

PFCR datasets are summarized in Table~\ref{table:pfcd_datasets}. Our dataset has the highest number of fact-checked claims. It also has the second-highest number of input claims and pairs after \textit{CrowdChecked}, but that dataset is significantly noisier. Finally, our dataset has by far the most languages, while the second biggest dataset in this regard has 5 language with only 50 samples per language.


\begin{table}
    \centering
\resizebox{\columnwidth}{!}{%
\begin{tabular}{lrrrr}
\toprule
 & Input claims & FC claims & Pairs & Languages \\ \midrule
 \citealp{kazemi_claim_2021} & \textit{NA} & ~150,000 & ~258 & 5 \\
\citealp{jiang_categorising_2021}             & \textit{NA} & 90 & 1,573 & 1 \\
\citealp{shaar-etal-2020-known}        & \textit{NA}  & 27,032 & 1,768 & 1\\ 
\citealp{checkthat_2021}         & 2,259 & 44,164 & 2,440 & 2\\
\citealp{crowdchecked_2022}        & 316,564 & 10,340 & 332,660 & 1\\ 
\midrule
\textit{MultiClaim} (our) & {\socialMediaPostsTotal} & {\factChecksTotal} & {\confirmedPairsTotal} & {\languagesTotal} \\ \bottomrule
\end{tabular}}
\caption{PFCR datasets. FC claims are \textit{fact-checked}. \textit{NA} means that we were not able to identify the correct number of input claims. The number should be similar to the number of pairs in most cases.}
\label{table:pfcd_datasets}
\end{table}

\paragraph{Methods.}
Methods used for PFCR are usually either BM25 (and other similar information retrieval algorithms) or various text embedding-based approaches~\cite[][i.a.]{voRiseGuardiansFactchecking2018, shaar-etal-2022-role, shaar_assisting_2021}. Reranking is often used to combine several methods to side-step compute requirements or as a sort of ensembling~\cite[][i.a.]{shaar-etal-2020-known}. PFCR task is also a target of the \textit{CLEF's CheckThat!} challenge, with many teams contributing with their solutions~\citep{checkthat_2022}. Other methods use visual information from images~\citep{mansour_did_2022, vo_lee_facts}, abstractive summarization~\citep{bhatnagar_abstractive_summarization_2022}, or key sentence identification~\citep{Sheng_2021} to improve the results.

\section{Our Dataset}
\label{sec:dataset}

Our dataset \textit{MultiClaim} consists of fact-checks, social media posts and pairings between them.

\paragraph{Fact-checks.}
We have collected the majority of fact-checks listed in the Google Fact Check Explorer, as well as fact-checks from additional manually identified major sources (e.g., Snopes) that were missing. Overall, we have collected {\factChecksTotal} fact-checks from {\factCheckingOrganizationsTotal} fact-checking organizations covering {\factChecksLanguages} languages. We publish the \textit{claim, title, publication date, and URL} of each fact-check. We do not publish the full body of the articles. The claim is usually (in 88.2\% of the cases) a one sentence long summarization of the information being fact-checked.

\paragraph{Social media posts.}
We used two ways to find relevant social media posts from Facebook, Instagram and Twitter. In both cases, it was professional fact-checkers that assigned the fact-checks to the posts. \textbf{(1)}~Some fact-checks use the \textit{ClaimReview} schema\footnote{\url{https://schema.org/ClaimReview}}, which has a field for reviewed items. All the links to the three social media platforms from this field are used to collect the posts and form the \textit{pairs}. \textbf{(2)}~We searched for URLs to Facebook and Instagram in the main body of the fact-checks. This is our pool of potentially relevant posts. Then we use the fact-checking warnings these two platforms provide. These warnings contain links to relevant fact-checking articles. We use these links to establish additional \textit{pairs}.

In total, we collected {\socialMediaPostsTotal} posts from {\socialMediaPostsLanguages} languages. There are {\confirmedPairsTotal} \textit{fact-check-to-post pairs}, each post in our dataset is paired with at least one fact-check. {\confirmedPairsInSameLanguage} of these pairs are monolingual and {\confirmedPairsInDifferentLanguage} are crosslingual (as predicted by the language identification, see below). Figure~\ref{fig:map} shows the major (more than 100 samples) languages. All the crosslingual cases have the visualized language for posts and English for fact-checks. We can see that there is a clear distinction between these two groups, probably caused by different fact-checking cultures in different regions.

\begin{figure}[t]
\centering
\includegraphics[width=1\columnwidth]{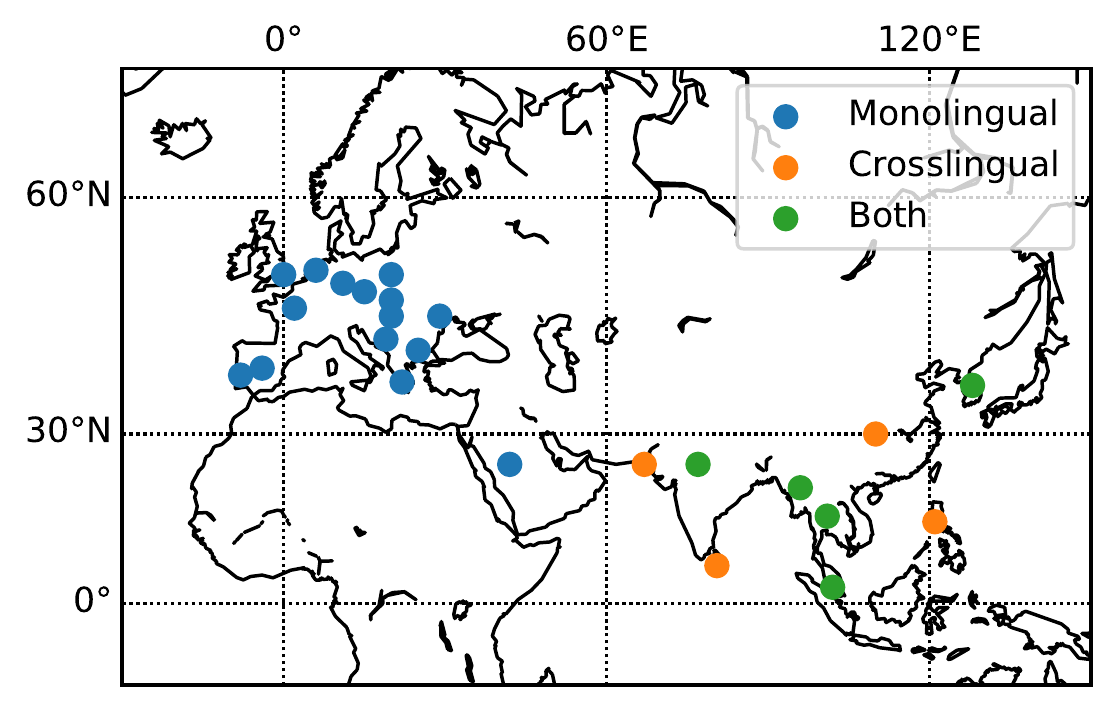}
\caption{Major languages from our dataset. Crosslingual languages all have English fact-checks.}\label{fig:map}
\end{figure}

We publish the \textit{text, OCR of the attached images (if any), publication date, social media platform, and fact-checker's rating} of each post. The \textit{rating} is the reason why the post was flagged (see Section~\ref{sec:dimensions} for more details). We do not publish URLs in an effort to protect the users and their privacy as much as possible. For detailed information about the implementation of this dataset collection pipeline, see Appendix~\ref{sec:dataset-collection-pipeline}. For a more detailed breakdown of dataset statistics (by languages and sources), see Appendix~\ref{sec:dataset-statistics}. Examples from our dataset can be seen in Appendix~\ref{app:examples}. 

\paragraph{Dataset versions.} We machine-translated all the published texts into English, resulting in two parallel versions of our dataset: the \textit{original version} and the \textit{English version}. We also used automatic language identification on all the texts. Both translations and language identifications are published as well.

\paragraph{Noise ratio.}
We manually checked 100 randomly selected pairs from our dataset and evaluated their validity. Three authors rated these pairs and assessed whether the claim from the fact-check was made in the post. In case of disagreement, they discussed the annotation until an agreement was reached. Based on our assessment, 87 out of 100 pairs were correct. The remaining 13 pairs were not errors made by social media platforms or fact-checkers, but rather posts that required visual information (either from video or image) to fully match the assigned fact-check. The 95\% Agresti-Coull confidence interval~\cite{agresti1998approximate} for correct samples in our dataset is 79-92\%.

\section{Unsupervised Evaluation}
\label{sec:unsupervised}

We formulate the task we are solving with our dataset as a ranking task, i.e., for each post, the methods rank all the fact-checks. Then, we evaluate the performance based on the rank of the desired fact-checks by using success-at-K (S@K) as the main evaluation metric. We define as the percentage of pairs when the desired fact-check ends up in the top K. Throughout the paper, we report this metric with the $95\%$ Agresti-Coull confidence interval.

For unsupervised evaluation, we evaluated text embedding models and the BM25 algorithm to understand how they are able to handle pairs in different languages or even crosslingual pairs. Fact-checks are represented with their claims only. Posts are represented with their main texts concatenated with the OCR transcripts. We use either the original texts or their English translations, depending on the version of the dataset that is reported.

\paragraph{Text embedding models (TEMs).} We use various neural TEMs~\cite{sbert_2019} that encode texts into a vector space. These are usually based on pre-trained transformer language models fine-tuned as Siamese networks to generate well-performing text embeddings. We use these models to embed both social media posts and fact-checked claims into a common vector space. The retrieval is then reduced to calculating and sorting distances between vectors.

\paragraph{BM25.} With BM25~\cite{bm25_2009}, we use the posts as queries and fact-checked claims as documents. The score is then calculated based on the lexical overlap between the query and all the documents.

\subsection{Main Results}\label{sec:main}

We compare the performance of 15 English TEMs, 5 multilingual TEMs, and BM25. The English TEMs were only evaluated with the \textit{English} version. The multilingual TEMs and BM25 were evaluated with both the \textit{original} and the \textit{English} versions. BM25 with different versions is denoted as BM25-Original and BM25-English, respectively.

In this section, we use different strategies to evaluate monolingual and crosslingual pairs. For monolingual pairs, we only search within the pool of fact-checks written in the same language as the post (e.g., for a French post we only rank the French fact-checks). For crosslingual pairs, we search in all the fact-checks\footnote{The index created for BM25 is multilingual as well.}. In both cases, we report the average performance for individual languages. We only report for languages with more than 100 pairs. For crosslingual pairs, we also consider a separate \textit{Other} category for all the leftover pairs.

\begin{figure}[t]
\centering
\includegraphics[width=0.9\columnwidth]{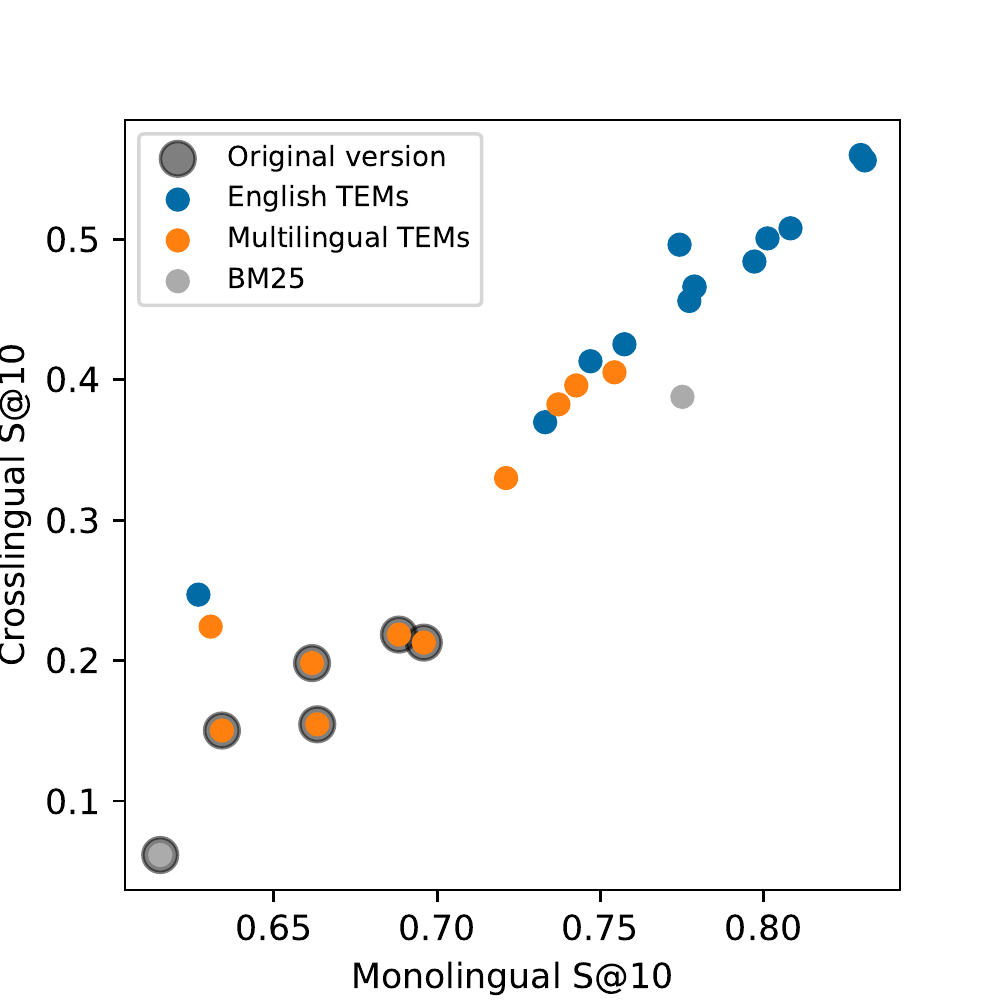}
\caption{Comparison of different method families. Unless stated otherwise, the methods use \textit{English} version of our dataset.}\label{fig:main}
\end{figure}

We present the main results in Table~\ref{tab:main} and we visualize them in Figure~\ref{fig:main}. We conclude that: \textbf{(1)} English TEMs are the best performing option for both monolingual and crosslingual claim retrieval. \textbf{(2)} Machine translation significantly improved the performance of both BM25 and TEMs. The difference between the best performing \textit{English} version method and the best performing \textit{original} version method is 35\% for crosslingual and 14\% for monolingual S@10. Currently, machine translation systems also have better language coverage than multilingual TEMs. \textbf{(3)} TEMs have a strong correlation between monolingual and crosslingual performance (Pearson's $\rho = 0.98$,  $P = 4{e-}10$ for English TEMs). These two capabilities do not conflict. \textbf{(4)} There is almost no correlation (Pearson's $\rho = 0.03$, $P = 0.89$ for English TEMs) between model size and performance. The training procedure is much more important. GTR is an exceptionally well-performing family, with all three models being Pareto optimal w.r.t. model size and performance. Another notable model is MiniLM -- a surprisingly powerful model for its size (33M).

\begin{table}[t]
    \centering
    \tiny
\begin{tabular}{lllrrr}
\toprule
Method & Size [M] & Ver. & Mono & Cross & SLB \\
\midrule
\textbf{BM25} \\
 &  & En & 0.78 & 0.39 & 0.18 \\
 &  & Og & 0.62 & 0.06 & 0.69 \\
\midrule
\textbf{English TEMs} \\
DistilRoBERTa & 82 & En & 0.76 & 0.43 & 0.18 \\
GTR-T5-Base & 110 & En & 0.81 & 0.51 & 0.19 \\
GTR-T5-Large & 336 & En & \textbf{0.83} & \textbf{0.56} & 0.20 \\
GTR-T5-XL & 1242 & En & \textbf{0.83} & \textbf{0.56} & 0.20 \\
MPNet-Base & 109 & En & 0.78 & 0.47 & 0.18 \\
MSMARCO-BERT-Base & 109 & En & 0.78 & 0.46 & 0.18 \\
MiniLM-L12 & 33 & En & 0.80 & 0.48 & 0.18 \\
MultiQA-MPNet-Base & 109 & En & 0.80 & 0.50 & 0.18 \\
SGPT-125M & 125 & En & 0.63 & 0.25 & 0.14 \\
SGPT-2.7B & 2700 & En & 0.77 & 0.50 & 0.19 \\
Sentence-T5-Base & 110 & En & 0.73 & 0.37 & 0.14 \\
Sentence-T5-Large & 336 & En & 0.75 & 0.41 & 0.15 \\
Sentence-T5-XL & 1242 & En & 0.78 & 0.47 & 0.16 \\
\midrule
\textbf{Multilingual TEMs} \\
DistilUSE-Base-Multilingual & 135 & En & 0.74 & 0.40 & 0.15 \\
 & & Og & 0.66 & 0.20 & 0.16 \\
LaBSE & 472 & En & 0.63 & 0.22 & 0.13 \\
 & & Og & 0.69 & \textbf{0.22} & 0.17 \\
MPNet-Base-Multilingual & 278 & En & 0.75 & 0.41 & 0.16 \\
 & & Og & \textbf{0.70} & 0.21 & 0.17 \\
MiniLM-L2-Multilingual & 118 & En & 0.74 & 0.38 & 0.16 \\
 & & Og & 0.63 & 0.15 & 0.17 \\
XLM-R & 278 & En & 0.72 & 0.33 & 0.15 \\
 & & Og & 0.66 & 0.15 & 0.16 \\
\bottomrule
\end{tabular}

\caption{Results for methods showing both \textit{mono}lingual and \textit{cross}lingual S@10. Ver. denotes either the \textit{original} (Og) or the \textit{English} (En) version of our dataset. The best results for these two versions are bolded. SLB denotes \textit{same language bias}.}\label{tab:main}
\end{table}

\begin{figure*}[t]
\centering
\includegraphics[width=1\textwidth]{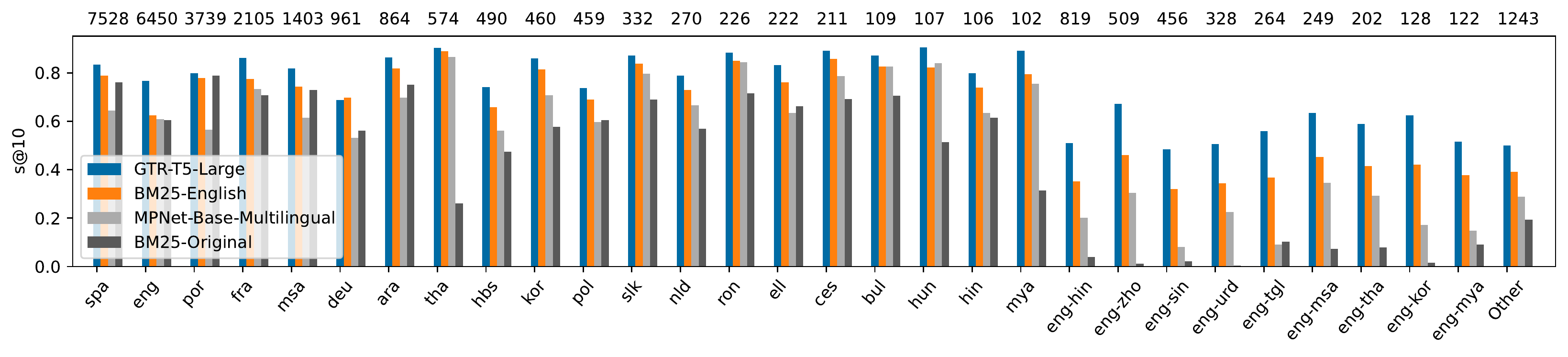}
\caption{Performance of selected methods for individual languages. For crosslingual pairs (e.g., \textit{eng-hin}), the first language is for the fact-checks and the second is for the posts. The number of pairs is shown at the top.}\label{fig:languages}
\end{figure*}

\paragraph{Languages.}
Performance for individual languages is shown in Figure~\ref{fig:languages}. We show the results for the best performing TEMs for both versions (\textbf{GTR-T5-Large} for the \textit{English} and \textbf{MPNet-Base-Multilingual} for the \textit{original}, which are denoted as GTR-T5 and MPNet from now on) and both BM25s. We cannot directly compare the performance numbers across different monolingual languages, since they use different pools of fact-checks with different sizes. This is also why smaller languages seem to have better scores.

BM25-Original, despite its seemingly weak overall performance, is actually competitive in some languages, e.g., Spanish, Portuguese, or Malay. It is better than multilingual TEMs for 7 out of 20 monolingual cases. Its overall monolingual performance is significantly decreased by Thai and Myanmar, due to their use of \textit{scriptio continua}. On the other hand, unlike multilingual TEMs, BM25-Original is by design not capable of any crosslingual retrieval and the results are shown only for completeness. 

\paragraph{False positive rate.}
We noticed that BM25-Original seems to perform better for languages with larger fact-check pools. We conducted an experiment to measure how pool size affects the results. We randomly selected 100 pairs for 7 of our languages with the largest fact-check pools.  We then measured the performance for these 100 pairs while increasing the pool size from 100 to 2,100 by gradually adding random fact-checks.

\begin{figure}[t]
\centering
\includegraphics[width=1\columnwidth]{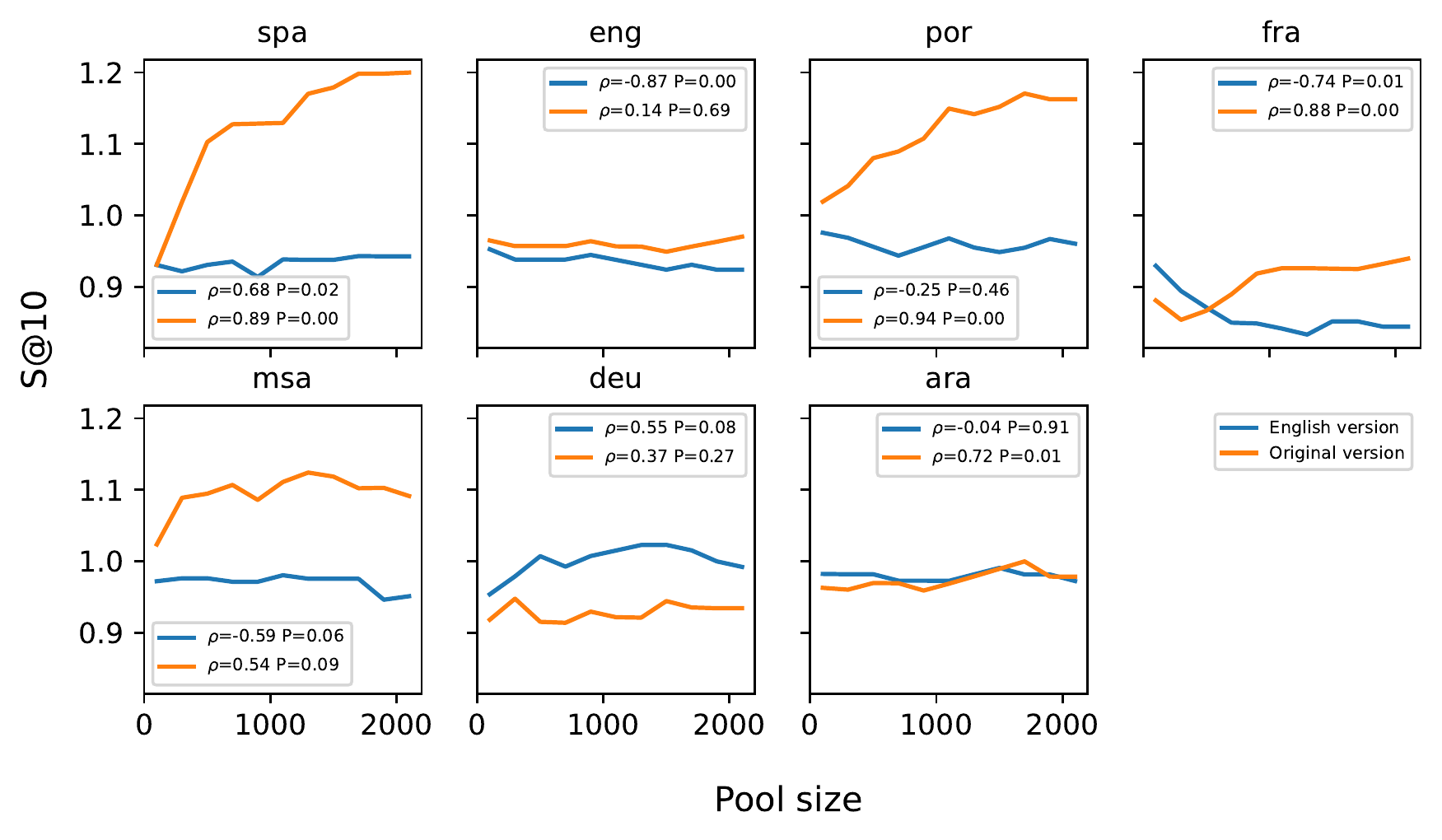}
\caption{Relative performance (S@10) between BM25 methods and TEMs for different fact-check pool sizes. For both versions we compare the best performing TEMs (GTR-T5 and MPNet) with BM25. Positive $\rho$ means that BM25 gets better with the growing pool size.}\label{fig:false_positive}
\end{figure}

We found that our initial observation was correct and that BM25-Original performs better than the MPNet model as the pool size increases (especially for Spanish, Portuguese, and French). The relative comparison between BM25 methods and TEMs is shown in Figure~\ref{fig:false_positive}. This suggests, that MPNet has a higher \textit{false positive rate}, i.e., it is more likely to assign high scores to irrelevant fact-checks. As the number of fact-checks grows, the risk of selecting irrelevant fact-checks also grows. \textbf{Different methods may be appropriate for different languages based on the number of fact-checks available.} We did not find the same pattern when comparing the methods using the \textit{English} version.

\paragraph{Same language bias.}
The fact that we reduce the fact-checks pool to one language in monolingual evaluation is motivated by what we call \textit{same language bias} (SLB) -- a tendency of methods to retrieve fact-checks that have the same language as the post. We approximate SLB by calculating the percentage of top 10 fact-checks that have the same language as the input post when we use the full pool. This number is reported in Table~\ref{tab:main}.

BM25-Original has the highest SLB score of all the methods, as it has an implicit language filter that effectively removes fact-checks from other languages from the pool. This reduction makes the task easier, but it violates our requirement that the method should take fact-checks in all the languages into consideration. We used language-filtered fact-checks in monolingual evaluation to reduce the effect the SLB has on the results. Without this filtering, BM25-Original would clearly outperform MPNet (S@10 $51.9$ vs $38.5$), even though our results in Figure~\ref{fig:languages} show that for many languages, its language understanding capabilities are actually worse.

\begin{figure}[t]
\centering
\includegraphics[width=1\columnwidth]{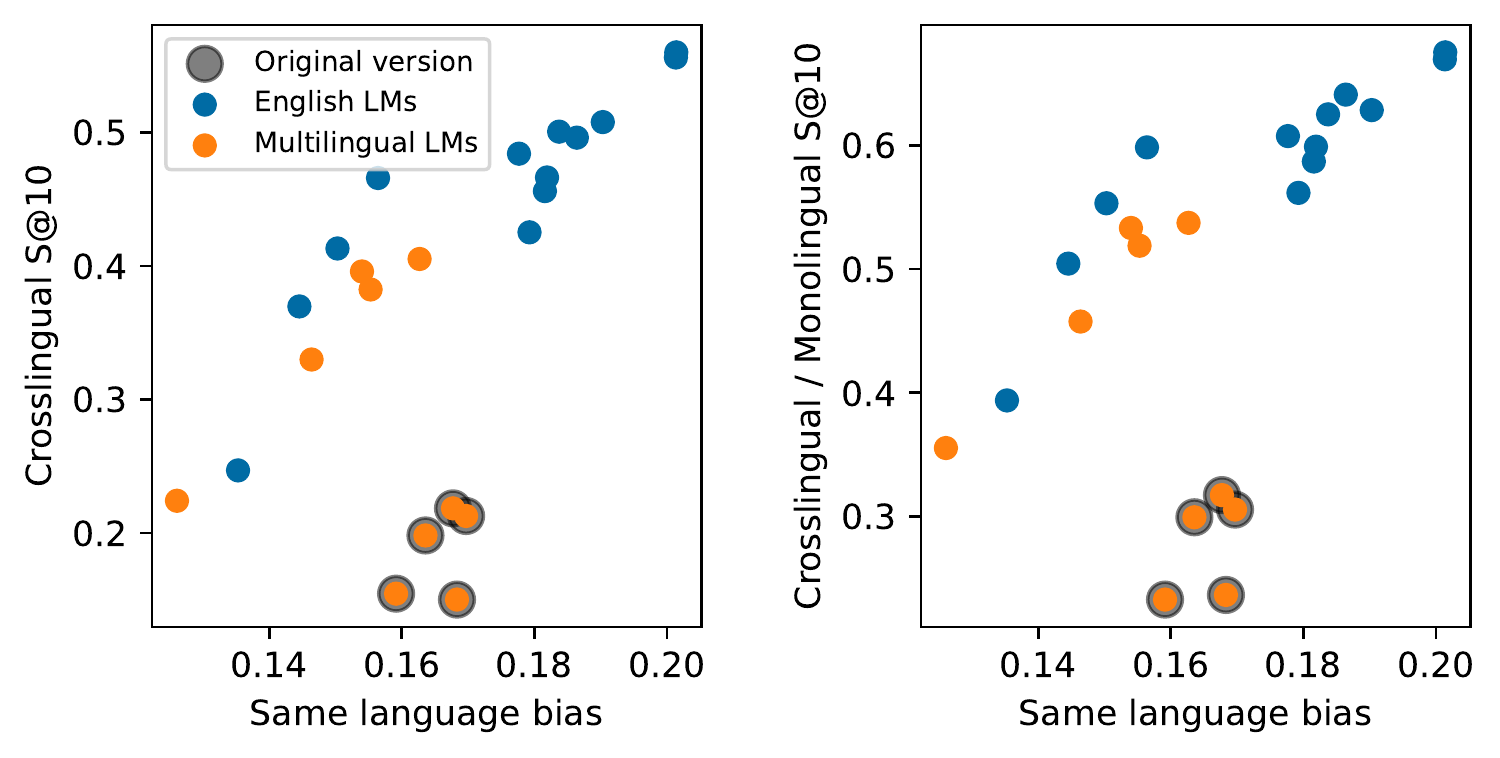}
\caption{Relation between \textit{same language bias} and performance for TEMs.}\label{fig:slb}
\end{figure}

However, it is not necessarily true that a higher SLB leads to worse crosslingual performance. As shown in Figure~\ref{fig:slb}, TEMs with the highest SLB actually have the best performance for crosslingual evaluation. Even more strikingly, the relative crosslingual performance compared to monolingual performance increases with SLB as well. We theorize that a certain amount of SLB is healthy, as long as the methods focus on meaningful similarities in texts written in the same language, such as local topics, named entities, and events, rather than on superfluous lexical overlaps. SLB can also be useful to localize claims that are not specific enough. For example, it is impossible to identify the country of origin for the following claim translated to English: \textit{Educational institutions are reopening from January 18}. However, as soon as we know that the original language was Bengali, we can guess that it is about Bangladeshi institutions.

\subsection{Other Dimensions}\label{sec:dimensions}

In this section, we report results for various data splits. Since we often work with small splits, we are not able to report the results as an average per language as in the previous section. Instead, we report the average score across the samples. This will give more weight to the more common languages, penalizing the methods with high \textit{false positive rate} (e.g., multilingual TEMs).

\begin{figure}[t]
\centering
\includegraphics[width=1\columnwidth]{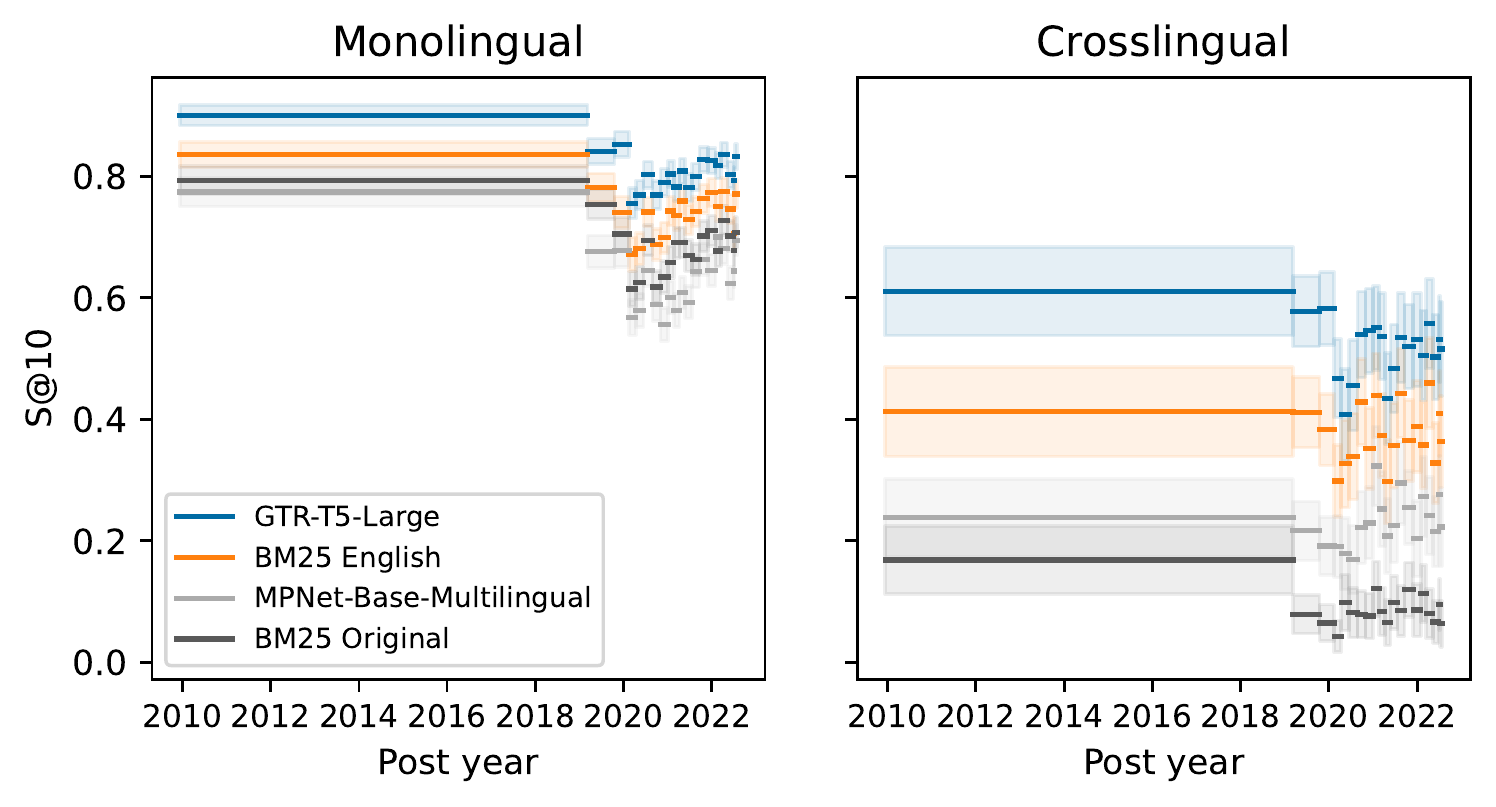}
\caption{Performance of selected methods for posts from different time intervals. Shaded areas are confidence intervals.}\label{fig:time}
\end{figure}

\paragraph{Time.}
We grouped the posts for which we were able to obtain the publication date (N = 26,337) into 20-quantiles and measured the performance of individual methods. The results are shown in Figure~\ref{fig:time}. There is a visible drop-off for all the methods at the start of 2020, largely caused by the COVID-19 pandemic. We confirmed this by measuring how well the methods worked on posts with the substrings \textit{corona}, \textit{covid} or \textit{korona}.\footnote{We chose this as a very simple, high-precision filtering technique. Many other COVID-19-related posts were not retrieved.} The results are shown in Table~\ref{tab:splits} (top panel).

The relative differences between individual methods seem stable. We hypothesized that TEMs might have problems with aging, since many of the foundation language models were originally trained before 2020. We correlated the average post time for each quantile with the difference between GTR-T5 and BM25-English performance and found a negative, but statistically insignificant correlation (Pearson $\rho = -0.33$, $P = 0.17$ for monolingual S@10). Similar results were measured for crosslingual performance. In both cases, the direction signals that the GTR model is indeed getting worse over time. We found no such signal comparing methods using the \textit{original} version.

There is a risk that the fact-check was written based on the very post we are using, and an information leak might have happened (e.g., the fact-checker might have used parts of the post verbatim). To test this, we compared pairs where the post is newer with the pairs where the post is older. We found that the two groups have virtually the same performance for all the methods (e.g., 80.02 vs 80.04 monolingual S@10 for GTR-T5). If there is an information leak happening, we were not able to measure it.

\paragraph{Post rating.}
In the case of Facebook and Instagram posts, fact-checkers use the so-called \textit{ratings} to describe the type of fallacy present. We show the results for the most common ratings in Table~\ref{tab:splits} (middle panel). \textit{Missing context} has a slightly lower score than \textit{(Partially) False information}. This might be caused by the fact that the rating is defined by what is \textit{not} written in the post, making it harder to match with an appropriate fact-check. \textit{Altered photo / video} rating has an even lower score. This is an expected behavior, since our purely text-based models cannot handle cases when the crux of the post is in its visual aspect.


\begin{table*}[t]
    \centering
    \tiny
\begin{tabular}{l|rllll|rllll}
\toprule
& \multicolumn{5}{c}{Monolingual} & \multicolumn{5}{c}{Crosslingual} \\
& $N$ & GTR-T5 & BM25-En & MPNet & BM25-Og & $N$ & GTR-T5 & BM25-En & MPNet & BM25-Og \\
\midrule
COVID-related & 4159 & 0.72 ± 0.01 & 0.68 ± 0.01 & 0.50 ± 0.02 & 0.60 ± 0.01 & 514 & 0.40 ± 0.04 & 0.29 ± 0.04 & 0.17 ± 0.03 & 0.06 ± 0.02 \\
Otherwise & 22615 & 0.83 ± 0.00 & 0.75 ± 0.01 & 0.66 ± 0.01 & 0.70 ± 0.01 & 3698 & 0.55 ± 0.02 & 0.39 ± 0.02 & 0.23 ± 0.01 & 0.08 ± 0.01 \\
\midrule
False information & 14812 & 0.82 ± 0.01 & 0.75 ± 0.01 & 0.65 ± 0.01 & 0.69 ± 0.01 & 2155 & 0.52 ± 0.02 & 0.37 ± 0.02 & 0.22 ± 0.02 & 0.09 ± 0.01 \\
Partly false information & 4498 & 0.82 ± 0.01 & 0.75 ± 0.01 & 0.63 ± 0.01 & 0.70 ± 0.01 & 669 & 0.53 ± 0.04 & 0.39 ± 0.04 & 0.21 ± 0.03 & 0.08 ± 0.02 \\
Missing context & 1993 & 0.77 ± 0.02 & 0.70 ± 0.02 & 0.61 ± 0.02 & 0.63 ± 0.02 & 268 & 0.53 ± 0.06 & 0.35 ± 0.06 & 0.19 ± 0.05 & 0.05 ± 0.03 \\
Altered photo/video & 753 & 0.73 ± 0.03 & 0.66 ± 0.03 & 0.52 ± 0.04 & 0.64 ± 0.03 & 142 & 0.47 ± 0.08 & 0.34 ± 0.08 & 0.17 ± 0.06 & 0.12 ± 0.05 \\
\midrule
Facebook & 24668 & 0.81 ± 0.00 & 0.74 ± 0.01 & 0.64 ± 0.01 & 0.68 ± 0.01 & 3927 & 0.52 ± 0.02 & 0.37 ± 0.02 & 0.22 ± 0.01 & 0.08 ± 0.01 \\
Instagram & 1473 & 0.78 ± 0.02 & 0.74 ± 0.02 & 0.56 ± 0.03 & 0.75 ± 0.02 & 44 & 0.56 ± 0.14 & 0.37 ± 0.14 & 0.19 ± 0.11 & 0.19 ± 0.11 \\
Twitter & 682 & 0.84 ± 0.03 & 0.74 ± 0.03 & 0.69 ± 0.03 & 0.70 ± 0.03 & 244 & 0.64 ± 0.06 & 0.49 ± 0.06 & 0.38 ± 0.06 & 0.06 ± 0.03 \\
\midrule
Total & 26774 & 0.81 ± 0.00 & 0.74 ± 0.01 & 0.64 ± 0.01 & 0.68 ± 0.01 & 4212 & 0.53 ± 0.02 & 0.38 ± 0.01 & 0.23 ± 0.01 & 0.08 ± 0.01 \\
\bottomrule
\end{tabular}
    \caption{Performance (S@10) with confidence intervals for various splits and methods.}
    \label{tab:splits}
\end{table*}

\begin{figure}[t]
\centering
\includegraphics[width=1\columnwidth]{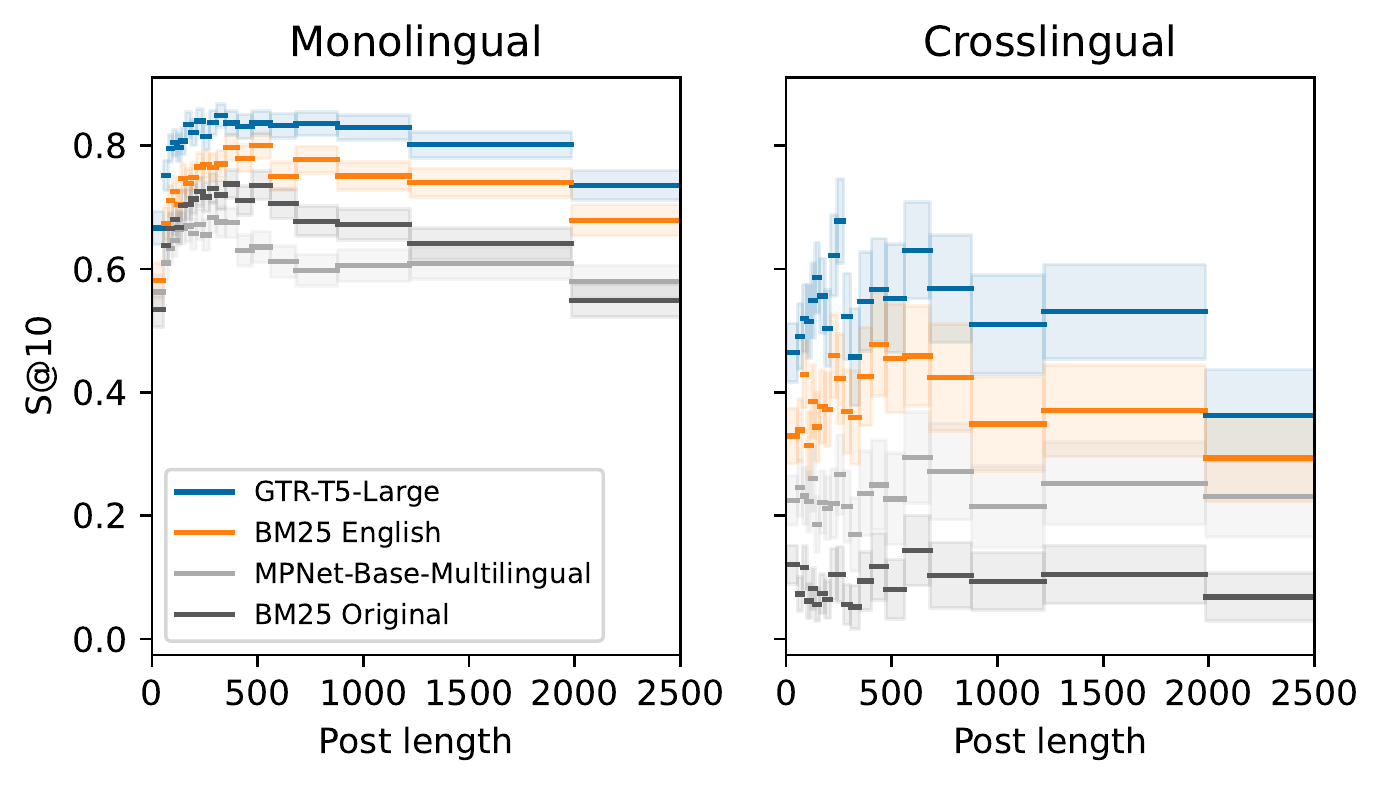}
\caption{Performance of selected methods for posts with different lengths. Shaded areas are confidence intervals.}\label{fig:length}
\end{figure}

\paragraph{Post length.} 
We show how the length of the posts influence the results in Figure~\ref{fig:length}. In general, the performance peaks at around 500 characters. Posts that are too short are too difficult to match (and extremely short posts may even indicate noise in the data). On the other hand, for posts longer than 500 characters, the methods gradually lose their effectiveness. The relative performance of methods seems to be relatively stable.

\paragraph{Social media platforms.} The results for social media platforms are in Table~\ref{tab:splits} (bottom pannel). We can see that Twitter has the best performance overall. We believe that this is, to a large extent, caused by the limited length of the Twitter posts.

\section{Supervised Training}
\label{sec:supervised}

To validate that our dataset can be used as a training set, we fine-tuned TEMs and evaluated their performance. We split the posts randomly into 80:10:10\% train, development, and test sets. We used \textit{cosine} or \textit{contrastive} training losses to fine-tune the models. In both cases, both positive and negative pairs are required for training. We used our data as positive samples and random pairs as negative samples. We performed a hyperparameter search with GTR-T5 and MPNet TEMs (see Section~\ref{sec:hparams_search}). Here, we report the best performing fine-tuned model we were able to achieve for both TEMs.

The overall results for the test set are reported in Table~\ref{tab:supervised}. We can see that GTR-T5 achieved only modest improvements\footnote{This is largely caused by the fact that we had to use smaller batch size due to (1) this model being larger, and (2) this model not supporting \textit{average mixed precision}. We believe that with larger batch size the performance could be even better.}. On the other hand, MPNet improved significantly in both monolingual and crosslingual performance, even surpassing the performance of BM25-English. We observed that the improvements were global across all languages.

We also observed that the TEMs were able to saturate the training set quite quickly, achieving 99.5\%+ average precision after only a few epochs. This shows that our naive random selection of negative samples was too easy. The model can learn only a limited amount of information from such samples, and we would need a more elaborate scheme for generating more challenging negative samples. This could lead to further performance improvements.

\begin{table*}[t]
    \centering
    \small
\begin{tabular}{l|rr|rr|rr}
\toprule
& \multicolumn{2}{c|}{Section~\ref{sec:supervised} (S@10)} & \multicolumn{2}{c|}{Section~\ref{sec:posthoc} (S@10)} &\multicolumn{2}{c}{Section~\ref{sec:posthoc} (R@10)}\\
Model & Monolingual & Crosslingual & Our dataset & Annotated & Our dataset & Annotated \\
\midrule
\multicolumn{3}{l}{\textbf{Unsupervised}} \\
GTR-T5-Large & 0.82 ± 0.01 & 0.55 ± 0.05 & 0.70 ± 0.09 & 0.93 ± 0.05 & 0.69 ± 0.09 & 0.59 ± 0.04 \\
BM25-English & 0.74 ± 0.02 & 0.40 ± 0.05 & 0.67 ± 0.10 & 0.85 ± 0.07 & 0.67 ± 0.09 & 0.48 ± 0.04 \\
MPNet-Base-Multilingual & 0.63 ± 0.02 & 0.23 ± 0.04 & 0.51 ± 0.10 & 0.70 ± 0.09 & 0.47 ± 0.09 & 0.32 ± 0.03 \\
BM25 Original & 0.68 ± 0.02 & 0.09 ± 0.03 & 0.60 ± 0.10 & 0.71 ± 0.09 & 0.58 ± 0.09 & 0.26 ± 0.03 \\
\midrule
\multicolumn{3}{l}{\textbf{Supervised}} \\
GTR-T5-Large & 0.84 ± 0.01 & 0.59 ± 0.05 & 0.71 ± 0.09 & 0.92 ± 0.05 & 0.70 ± 0.09 & 0.65 ± 0.03 \\
MPNet-Base-Multilingual & 0.76 ± 0.02 & 0.42 ± 0.05 & 0.62 ± 0.10 & 0.85 ± 0.07 & 0.60 ± 0.09 & 0.45 ± 0.04 \\
\bottomrule
\end{tabular}
    \caption{Test set performance (Section~\ref{sec:supervised}) and annotated results performance (Section~\ref{sec:posthoc}) of unsupervised and supervised methods.}
    \label{tab:supervised}
\end{table*}

\section{Post-Hoc Results Analysis}\label{sec:posthoc}

The pairs, we obtained from the fact-checks, are only a subset of all the potentially valid pairs. This incompleteness limits our understanding of the dataset and also our evaluation. We decided to manually annotate a subset of the results generated by the methods to better understand what is missing from our data. We generated the top 10 fact-checks for the 87 test set posts that we knew had valid fact-checks (see Section~\ref{sec:dataset}). We used the 4 unsupervised and 2 supervised methods from Section~\ref{sec:supervised}.

These methods generated 3,390 unique pair predictions for these 87 posts. Three authors went through each prediction and marked, whether they agreed with it, i.e., whether they found the fact-check to be valid and useful for the post. The agreement rates between the annotators were sufficiently high: 82.2\%, 85.5\% and 92.9\%. We consider pairs where at least two annotators agreed to be \textit{correct}. In total, the methods were able to find 719 correct pairs. 96 of these were present in our original dataset. This suggests that there is roughly 7$\times$ more pairs in our dataset than we had previously identified. No method was able to find 9 fact-checks out of 105 that were already in our dataset. Of the 719 correct pairs, only 247 were monolingual, 136 were crosslingual with an English fact-check, and 336 were crosslingual with a non-English fact-check. The last category in particular is almost completely missing from our dataset. 

In Table~\ref{tab:supervised}, we show the results for individual methods. We compare S@10 (now defined as how many posts have at least one correct fact-check produced in the top 10) as approximated with our dataset and the true S@10 obtained by the annotation. We can see that the score for our dataset is significantly lower and the true performance of our methods is better then what was measured previously. We also compare recall-at-10 (R@10), defined as the percentage of expected pairs a method was able to produce in the top 10. In this case, both our dataset and manual annotation are only estimates, since they do not contain \textit{all} the valid pairs, they both contain only a subset obtained by different methods. Here we can see that our dataset actually provides higher estimates. We assume that our annotation is more precise, so we conclude that  the recall calculated from our dataset is overinflated (possibly due to selection bias). \textbf{It also seems that our dataset has a bias in favor of BM25}, compared to the results obtained from annotated data.

\section{Discussion}

\paragraph{Complexity of crosslingual evaluation.}
Phenomena such as \textit{same language bias} or \textit{false positive rate} make the evaluation of multilingual and crosslingual datasets inherently complex. If we were to abstract the whole evaluation into a single number, as is often done in practice, we would have completely missed these pitfalls. Without an in-depth evaluation, we might have been misled while applying our methods in practice, e.g., while developing helpful tools for fact-checkers. Our evaluation procedures were previously impossible to develop in the absence of linguistically diverse PFCR datasets. 

\paragraph{Machine translation beats multilingual TEMs.}
These two technologies represent the two main multilingual and crosslingual learning paradigms -- label transfer and parameter transfer~\cite{pikuliak2021cross}. Machine translation is a clear winner in our case. English TEMs significantly outperform multilingual approaches for both monolingual and crosslingual retrieval.

\paragraph{COVID-19.}
As shown in Table~\ref{tab:splits}, it seems that the performance for COVID-19 is significantly worse than for the rest of the dataset. However, this might not necessarily mean that the methods are having issues with the domain shift. The sheer amount of fact-checks written about COVID-19 makes it hard for the methods to pick the desired fact-check in the presence of thousands of other very similar ones. This is evident considering that BM25 also has worse results, even though it should be less prone to domain shift based on its design.

\section{Conclusions}
\label{sec:conclusions}

In this paper, we introduced a new multilingual \textit{previously fact-checked claim retrieval} dataset. Our collection process yielded a unique and diverse dataset with a relatively small amount of noise in it. We believe that the evaluation of various methods is also insightful and can lead to the development of better fact-checking tools in the future.

We believe that our dataset opens up many interesting research directions. We have, for example, barely scraped the surface of crosslingual learning in this work. Applying various transfer learning methods (especially for low-resource languages) is an important future direction. 


\section{Limitations}
\label{sec:limitations}

\subsection{Dataset}

\paragraph{Noise.} Based on our annotation (see Section~\ref{sec:dataset}), we expect that $\sim$ 13\% of posts in our dataset do not contain the claim in the textual form. These are the cases when the claim being made on the social media is based on visual information. Note, that the methods might still be able to retrieve correct fact-checks for some of these posts, based on spurious correlations, e.g., overlaps in named entities.
    
\paragraph{AI APIs.} We use out-of-the-box AI services to perform optical character recognition, machine translation to English and language detection. All of these have limited precision and might inject noise into our data.

\begin{itemize}
    \item \textit{OCR} was too sensitive and was often reading imaginary character, watermarks, etc. We had to address this by a more aggressive text cleaning.
    \item \textit{Machine translation to English} is not perfect and the quality of translations depends on source language, particular topics or even the writing style.
    \item \textit{Language detection} is an important component in our pipeline as we use it to group samples by language and then reason about these languages. Noise in language detection might have influenced our results and insights.
\end{itemize}

\paragraph{Selection bias.} There is a possibility that selection bias influences our results. First, sometimes fact-checkers writing the fact-checks base their writing on a particular post and the fact-check might contain parts of it verbatim. We tried to measure the size of this effect by comparing cases when the fact-checks are newer and older then posts (see Section~\ref{sec:dimensions}), but we did not find a signal that this is the case. However, we know that there are at least few samples with this problem.

Second, there might be a bias towards social media posts that the social media platform or fact-checkers are already able to detect. Other, more difficult cases might still elude us.

\paragraph{Linguistic bias.} Although our dataset is quite diverse, compared to most published datasets, there is still a bias towards major languages and Indo-European language family in particular. Crosslingual pairs consist mostly of East or South Asian posts with non-Latin script mapped to English fact-checks. It is hard to estimate how our results would generalize to other language pairs. We visualize the languages in Figure~\ref{fig:map}. The annotation efforts in Section~\ref{sec:posthoc} show that there are many crosslingual pairs that our data collection methodology was not able to collect.

\subsection{Methods}

\paragraph{Language support.} The methods we use have different degrees of support for different languages. BM25 requires a proper tokenization to work. We have languages that use \textit{scriptio continua} -- Thai and Myanmar -- where this is a problem. BM25-Original performance for these two is subpar, but could be improved by implementing custom tokenization models.

Multilingual TEMs we use do not support Sinhala and Tagalog languages, i.e., they were not trained with their data. The performance for these two languages is again subpar. Additionally, all methods depending on machine translation are naturally only able to handle languages that have a machine translation system available, although we believe that this was not a significant problem in our dataset.
    
\paragraph{Hidden positive pairs.} The results we report might be deflated from the practical point of view because of unmarked correct pairs that are in the dataset. We have information only about a small subset of all the pairs. Our attempt to approximate true performance is provided in Section~\ref{sec:posthoc}.

\paragraph{Supervised learning overfitting.} It is possible that our supervised training yielded model that is overfitted on the particular languages and time frame that are represented in our dataset. The increase in performance might not transfer to out-of-domain pairs.

\section{Ethical Considerations}
\label{sec:ethics}

We analyzed the likelihood and impact of ethical and societal risks for the most affected stakeholders, such as social media users and profile owners, fact-checkers, researchers, or social media platforms. For the most severe risks, we proposed respective countermeasures, following the guidelines and arguments in~\citep{aoir_v3_2020, townsend2016social, mancosu_what_2020}. 

\paragraph{Data collection process.} 
While Twitter posts were collected using a (at the time of collection) publicly available API, the Terms of Service (ToS) of Facebook and Instagram do not currently allow for the accessing or collecting of data using automated means. Following the discussion and arguments presented in~\citep{mancosu_what_2020} and 
to minimize the harm to these social media platforms and their users, we made sure to only collect publicly available posts that are accessible even without logging in. 
Even if we admit the risk that such research activities could potentially violate the ToS, we argue that ignoring posts from Facebook and Instagram would prohibit research that seeks to address key current issues such as disinformation on these platforms~\cite{bruns_after_2019}. These are some of the main platforms for disinformation dissemination in many countries. 
We consider the collection of such public data and its usage for research purposes to be in the public interest, especially considering the status of disinformation as a hybrid security threat~\citep{enisa_2022}, which could justify minor harms to social media platforms.

Other considerable risks include the risk of accessibility privacy intrusion~\citep{tavani_ethics_2016} of social media users by observing them in an environment where they do not want to be observed. We did not obtain explicit consent from social media users to collect their posts. However, the criteria for considering social media data private or public depend on the assumption of whether social media users can reasonably expect to be observed by strangers~\cite{townsend2016social}. Twitter is considered an open platform. The collected posts on Facebook or Instagram are not only public, but the users can also expect that their posts will be widely shared, commented or reacted to and they can end up being fact-checked if it is the case. 

\paragraph{Data publication.} To minimize the risk of third-party misuse, the dataset is available only to researchers for research purposes. The full texts of the fact-checks are not published to avoid possible copyright violations.

We assessed the risk of re-identification, as well as the risk of revealing incorrect, highly sensitive or offensive content regarding social media users. At the same time, we had to take into account the fact that social media platforms remove some posts after they have been flagged as disinformation. Therefore, we decided to include the original texts of the posts in the dataset to prevent it from decaying. Otherwise, it would become progressively less usable and research based on it less reproducible. This also allows us to avoid publishing the URLs of posts, which would directly reveal the identities of the users. It is not possible to guarantee complete anonymity, since the posts are still linked in the fact-checks. The posts could also theoretically be found by full-text search.

On the other hand, all the posts released in our dataset are already mentioned in a publicly available space in the context of fact-checking efforts. Our publication of these posts does not significantly increase their already existing public exposure, especially considering the limited access options of our dataset.

To support users' rights to rectification and erasure in case of the publication of incorrect or sensitive information, we provide a procedure for them to request the removal of their posts from the dataset. However, we assess that the risk of wrongfully assigned fact-checks has a low probability (see Section~\ref{sec:dataset}). 


As the dataset can also be used for supervised training (see Section~\ref{sec:supervised}), there is a risk of propagating biases present in the data (see Section~\ref{sec:limitations}). We recommend performing a proper linguistic analysis of any supervised model w.r.t. all the languages for which the model is intended. The results shown in this paper may not reflect the performance of the methods on other languages. We are also aware of the risk of propagating the biases of the fact-checkers, as it is they who decide what to fact-check. Although they should generally follow principles of fact-checking ethics (see, e.g., the IFCN's Code of Principles), there may still be present some human or systemic biases~\cite{schwartz2022_nist} 
that could affect the results when using the dataset for other purposes. 

\section*{Acknowledgements}

This work was partially supported by the \textit{Central European Digital Media Observatory (CEDMO)}, a project funded by the European Union under the Contract No. 2020-EU-IA-0267, and by \textit{DisAI - Improving scientific excellence and creativity in combating disinformation with artificial intelligence and language technologies}, a project funded by European Union under the Horizon Europe, GA No. \href{https://doi.org/10.3030/101079164}{101079164}.

We thank \textit{Weights \& Biases} for providing us a license for their experiment tracking platform.

\bibliography{anthology,references}
\bibliographystyle{acl_natbib}

\appendix

\section{Computational Resources}

We calculated all the results on an AWS-based virtual machine located in the Ohio AWS data center. The machine has one NVIDIA Tesla T4 GPU installed. The unsupervised experiments would take approximately 2 GPU days to replicate. The supervised experiments would take approximately 3 GPU days to replicate. Additional roughly 4 GPU days were spent on other experiments that were discarded or are not reported in this paper.

\section{Dataset Pipeline Details}
\label{sec:dataset-collection-pipeline}

\subsection{Dataset Collection}

The dataset was collected via our research platform \textit{Monant}~\cite{srba2019monant}.

\paragraph{Crawling.} We use a \textit{Selenium}-based web crawler that visits the links, extracts the HTML content and parses it with the \textit{Beautiful Soup}\footnote{\url{https://www.crummy.com/software/BeautifulSoup/}} library.

\paragraph{Source of fact-checks.} We only processed fact-checks written by the AFP news agency. We chose them because they are an established fact-checking organization with high editing standards and are also a part of Meta's \textit{Third-Party Fact-Checking Program}. Pairs with fact-checks from other organizations might have been established from the warnings.

\paragraph{Archiving services.} Since the content from social media networks may disappear in time, fact-checkers tend to use various content archiving services (e.g., \texttt{perma.cc}). We extract the content from these services as well.

\paragraph{AI APIs.} We use following services to process our samples:

\begin{itemize}
    \item \textit{Google Vision API.} We use Google Vision API to extract text from images attached to the post. The API also returns a list of languages found in each image with their percentage.

    \item \textit{Google Translate API.} We use Google Translate API to translate all the texts into English. The API also returns a most probable language.
\end{itemize}

\subsection{Dataset Pre-Processing}
\label{sec:pre-processing-details}

We performed several cleaning and pre-processing steps with our dataset. All the pre-processing is available in the released code repository.

\paragraph{Removing noisy claims.} We removed fact-checks that had no claim or where the claim was shorter than 10 characters.

\paragraph{Fact-check deduplication.} We unified fact-checks with identical claims.

\paragraph{Noise in social media posts.} We removed texts or OCR transcripts that we deemed noisy (shorter than 25 characters or more than 50\% non-alphabetical characters). We then only kept posts where at least one text was considered not noisy. We also removed noisy lines from OCR transcripts (Lines shorter than 5 characters or with more than 50\% non-alphabetical characters). We also removed URLs.

\paragraph{Post deduplication.} We unified posts that ended up with identical text contents after the cleaning process.

\paragraph{Machine translation.} We translated all the texts into English. The only exceptions were fact-check claims coming from English-language providers (e.g., Snopes) that we considered English by default, and fact-check claims where CLD3\footnote{\url{https://github.com/google/cld3}} identified English language. We confirmed experimentally that CLD3 has a high precision on English texts.

\paragraph{Language identification normalization.} We observed that there are some systematic errors in the language identification models we used. We found out that the model often selected less common languages based on spurious patterns, e.g., mentions of Filipino politicians sometimes led to Ilocano language prediction. Based on data analysis, we changed some predictions automatically, e.g., all Ilocano predictions were changed into English. Sometimes we only did it when the script did not match the language, e.g., for posts with Latin  script identified as Oromo. We do not recommend using this process automatically on any data. In other contexts, the generated predictions might be less noisy. Even in our case, we have different rules for posts and fact-checks based on the characteristics of these two domains. If the predictions proved to be too noisy, we unified several languages or language varieties into one. This is the case of Croatian, Bosnian and Serbian, as well as Indonesian and Malay.

\section{Dataset Statistics}
\label{sec:dataset-statistics}

We show the number of fact-checks and posts per language in Table~\ref{tab:language_stats}. For fact-checks, we only take into consideration the language of claim, since we mostly only work with claims in this work.

Posts can have more than one language detected based on its overall compositions. We calculated percentage for each language based on the language prediction methods. We consider all languages with at least 20\% to be relevant. 25,482 posts have only one language detected, while 2,549 has two, 59 has three, 1 has four and 1 has zero.

\begin{table}
    \centering
    \small
\begin{tabular}{llrr}
\toprule
Code &           Language &  \# fact-checks &  \# posts \\
\midrule
 ara &             Arabic &          14201 &     931 \\
 asm &           Assamese &             60 &       5 \\
 aze &        Azerbaijani &            178 &       2 \\
 bul &          Bulgarian &            162 &     114 \\
 ben &            Bengali &           4143 &     113 \\
 cat &            Catalan &            574 &     100 \\
 ces &              Czech &            254 &     265 \\
 dan &             Danish &            648 &       6 \\
 deu &             German &           4996 &     932 \\
 ell &              Greek &           1821 &     175 \\
 eng &            English &          85814 &    7307 \\
 spa &            Spanish &          14082 &    7319 \\
 fas &              Farsi &            418 &      17 \\
 fin &            Finnish &            109 &     103 \\
 tgl &            Tagalog &            462 &     439 \\
 fra &             French &           4355 &    2146 \\
 hbs &     Serbo-Croatian &           2451 &     481 \\
 hin &              Hindi &           7149 &     833 \\
 hun &          Hungarian &            139 &     113 \\
 ita &            Italian &           3047 &      65 \\
 heb &             Hebrew &            202 &       2 \\
 jpn &           Japanese &             62 &       7 \\
 khm &              Khmer &            144 &       6 \\
 kor &             Korean &            510 &     474 \\
 mkd &         Macedonian &           1125 &       1 \\
 mal &          Malayalam &           1206 &       4 \\
 msa &              Malay &           8424 &    1389 \\
 mya &            Myanmar &             92 &     172 \\
 nld &              Dutch &           1232 &     257 \\
 nor &          Norwegian &            440 &       5 \\
 pol &             Polish &           6912 &     453 \\
 por &         Portuguese &          21569 &    3366 \\
 ron &           Romanian &            204 &     238 \\
 rus &            Russian &           2715 &      28 \\
 sin &            Sinhala &            825 &     534 \\
 slk &             Slovak &            260 &     363 \\
 sqi &           Albanian &            726 &       1 \\
 tam &              Tamil &           1612 &      29 \\
 tel &             Telugu &           2450 &      11 \\
 tha &               Thai &            382 &     626 \\
 tur &            Turkish &           6676 &       7 \\
 ukr &          Ukrainian &             68 &       6 \\
 urd &               Urdu &              0 &     378 \\
 zho &            Chinese &           2586 &     595 \\
     &             Others &            266 &     343 \\
\bottomrule
\end{tabular}
\caption{List of languages with at least 50 fact-checks or 50 posts.}\label{tab:language_stats}
\end{table}

Table~\ref{tab:sources} shows the sources of our fact-checks. Here we only show the statistics for the fact-checks we actually used in our experiments. There are additional 6k fact-checks that we have not used because they we were not able to fill their \textit{claim} field.

\begin{table*}
    \centering
    \tiny
\begin{tabular}{llr|llr|llr}
\toprule
 Name & Lang. & $N$ & Name & Lang. & $N$ & Name & Lang. & $N$ \\
\midrule
snopes.com & eng & 18376 & washingtonpost.com & eng & 1413 & agi.it & ita & 246 \\
politifact.com & eng & 9029 & dogrulukpayi.com & tur & 1360 & verify-sy.com & ara & 242 \\
misbar.com & ara & 9027 & stopfake.org & rus & 1307 & cbsnews.com & eng & 242 \\
boomlive.in & eng & 7949 & colombiacheck.com & spa & 1271 & factchecknederland.afp.com & nld & 234 \\
factcheck.afp.com & eng & 6853 & tempo.co & id & 1143 & butac.it & ita & 220 \\
cekfakta.com & ind & 6523 & vistinomer.mk & mkd & 1141 & efe.com & spa & 219 \\
altnews.in & eng & 6199 & faktograf.hr & hbs & 1094 & br.de & deu & 214 \\
factly.in & eng & 5818 & dubawa.org & eng & 1066 & annielab.org & eng & 204 \\
leadstories.com & eng & 5319 & factcheck.kz & rus & 1044 & globes.co.il & heb & 202 \\
sapo.pt & por & 5200 & istinomer.rs & hbs & 958 & factcheckhub.com & eng & 200 \\
demagog.org.pl & pol & 4292 & boombd.com & ben & 937 & ghanafact.com & eng & 199 \\
fullfact.org & eng & 4260 & bufale.net & ita & 928 & telemundo.com & spa & 195 \\
factual.afp.com & spa & 4051 & apublica.org & por & 915 & apa.at & deu & 185 \\
uol.com.br & por & 3908 & rappler.com & tgl & 874 & verificat.afp.com & ron & 177 \\
checkyourfact.com & eng & 3620 & verificat.cat & spa & 821 & efectococuyo.com & spa & 170 \\
teyit.org & tur & 3289 & kallxo.com & sqi & 728 & factcheckni.org & eng & 157 \\
newsmobile.in & eng & 3265 & aap.com.au & eng & 687 & proveri.afp.com & bul & 152 \\
newtral.es & spa & 3256 & projetocomprova.com.br & por & 686 & icirnigeria.org & eng & 142 \\
dpa-factchecking.com & nld & 2839 & tjekdet.dk & dan & 648 & tenykerdes.afp.com & hun & 138 \\
indiatoday.in & eng & 2799 & dogrula.org & tur & 634 & liberation.fr & fra & 134 \\
factcheck.org & eng & 2716 & faktencheck.afp.com & deu & 629 & factcheckgreek.afp.com & ell & 129 \\
aosfatos.org & por & 2596 & thip.media & ben & 598 & radio-canada.ca & fra & 123 \\
boatos.org & por & 2553 & dailyo.in & ben & 591 & maharat-news.com & ara & 121 \\
aajtak.in & hin & 2493 & univision.com & spa & 582 & factcheckmyanmar.afp.com & mya & 119 \\
dabegad.com & ara & 2342 & periksafakta.afp.com & ind & 563 & jachai.org & ben & 113 \\
factcheck.afp.com/ar & ara & 2292 & lemonde.fr & fra & 558 & nieuwscheckers.nl & nld & 111 \\
estadao.com.br & por & 2197 & check4spam.com & eng & 524 & europapress.es & spa & 108 \\
factuel.afp.com & fra & 2178 & healthfeedback.org & eng & 499 & faktantarkistus.afp.com & fin & 107 \\
thequint.com & eng & 2058 & mygopen.com & zho & 494 & tagesschau.de & deu & 103 \\
tfc-taiwan.org.tw & zho & 1960 & sprawdzam.afp.com & pol & 458 & scroll.in & eng & 100 \\
observador.pt & por & 1930 & faktisk.no & nor & 444 & thelallantop.com & hin & 99 \\
usatoday.com & eng & 1901 & presseportal.de & deu & 439 & theferret.scot & eng & 96 \\
oko.press & pol & 1872 & 20minutes.fr & fra & 419 & france24.com & fra & 92 \\
fatabyyano.net & ara & 1844 & cinjenice.afp.com & hbs & 387 & voachinese.com & zho & 92 \\
factcrescendo.com & ben & 1808 & factcheckthailand.afp.com & tha & 382 & comprovem.afp.com & cat & 90 \\
correctiv.org & deu & 1783 & factcheckkorea.afp.com & kor & 382 & factandfurious.com & fra & 82 \\
maldita.es & spa & 1748 & asianetnews.com & mal & 365 & factchecker.in & eng & 74 \\
ellinikahoaxes.gr & ell & 1688 & newsweek.com & eng & 364 & telugupost.com & tel & 73 \\
checamos.afp.com & por & 1672 & factnameh.com & fas & 356 & zimfact.org & eng & 72 \\
facta.news & ita & 1652 & fakenews.pl & pol & 320 & factcheckbangla.afp.com & ben & 62 \\
youturn.in & tam & 1609 & fastcheck.cl & spa & 313 & buzzfeed.com & eng & 56 \\
malumatfurus.org & tur & 1572 & newsmeter.in & eng & 290 & verificado.com.mx & spa & 55 \\
polygraph.info & eng & 1527 & factrakers.org & eng & 276 & ripplesnigeria.com & eng & 52 \\
metafact.io & eng & 1526 & semakanfakta.afp.com & msa & 267 & poynter.org & eng & 52 \\
africacheck.org & eng & 1468 & fakty.afp.com & slk & 260 & globo.com & por & 52 \\
animalpolitico.com & spa & 1468 & napravoumiru.afp.com & ces & 255 & radiofarda.com & fas & 51 \\
verafiles.org & tgl & 1414 & factograph.info & rus & 253 & stern.de & deu & 50 \\

\bottomrule
\end{tabular}
\caption{Fact-checking sources with at least 50 fact-checks in our dataset.}\label{tab:sources}
\end{table*}

Table~\ref{tab:language_matrix} show the number of fact-check-to-post pairs for different language combinations. 

\begin{table*}
\tiny
\centering
\setlength{\tabcolsep}{3pt} 
    \begin{tabular}{ll|rrrrrrrrrrrrrrrrrrrrrrrrr}
    \toprule
    & & \multicolumn{25}{c}{\textit{Fact-check language}} \\
 & & ara & ben & bul & cat & ces & deu & ell & eng & fin & fra & hbs & hin & hun & kor & msa & mya & nld & pol & por & ron & sin & slk & spa & tha & Other \\ \midrule
\multirow{28}{*}{\rotatebox[origin=c]{90}{\textit{Social media post language}}} & ara & 864 & 0 & 1 & 0 & 0 & 0 & 0 & 28 & 1 & 0 & 23 & 0 & 0 & 0 & 0 & 3 & 0 & 0 & 0 & 0 & 0 & 0 & 0 & 0 & 0 \\
& ben & 0 & 109 & 0 & 0 & 0 & 0 & 1 & 0 & 0 & 0 & 0 & 1 & 0 & 0 & 0 & 0 & 0 & 0 & 0 & 0 & 0 & 0 & 0 & 0 & 1 \\
& bul & 0 & 0 & 56 & 0 & 0 & 0 & 0 & 75 & 0 & 0 & 0 & 0 & 1 & 0 & 0 & 1 & 0 & 0 & 0 & 0 & 0 & 0 & 0 & 0 & 0 \\
& cat & 0 & 0 & 0 & 79 & 0 & 0 & 0 & 0 & 10 & 0 & 0 & 0 & 0 & 0 & 0 & 1 & 0 & 0 & 0 & 0 & 0 & 0 & 0 & 0 & 0 \\
& ces & 0 & 0 & 0 & 0 & 211 & 2 & 1 & 5 & 1 & 0 & 0 & 2 & 0 & 0 & 0 & 0 & 0 & 0 & 1 & 0 & 0 & 0 & 28 & 0 & 0 \\
& deu & 1 & 0 & 0 & 0 & 1 & 961 & 1 & 5 & 1 & 0 & 0 & 4 & 0 & 0 & 0 & 0 & 0 & 0 & 2 & 1 & 0 & 0 & 3 & 1 & 1 \\
& ell & 0 & 0 & 0 & 0 & 0 & 0 & 222 & 3 & 0 & 0 & 0 & 0 & 0 & 0 & 0 & 0 & 0 & 0 & 0 & 0 & 0 & 0 & 0 & 0 & 0 \\
& eng & 2 & 1 & 2 & 4 & 12 & 37 & 4 & 6450 & 77 & 5 & 50 & 8 & 16 & 3 & 25 & 67 & 0 & 12 & 20 & 34 & 3 & 5 & 14 & 12 & 28 \\
& fin & 3 & 0 & 0 & 43 & 0 & 1 & 2 & 39 & 7528 & 0 & 5 & 0 & 0 & 0 & 0 & 4 & 0 & 0 & 6 & 28 & 1 & 0 & 1 & 0 & 13 \\
& fra & 0 & 0 & 0 & 0 & 0 & 3 & 0 & 0 & 0 & 95 & 1 & 0 & 0 & 0 & 0 & 0 & 0 & 0 & 0 & 2 & 0 & 0 & 0 & 0 & 0 \\
& hbs & 0 & 0 & 0 & 0 & 0 & 0 & 0 & 264 & 0 & 0 & 0 & 0 & 0 & 0 & 0 & 2 & 0 & 0 & 0 & 0 & 0 & 0 & 0 & 0 & 0 \\
& hin & 1 & 0 & 3 & 0 & 0 & 1 & 2 & 20 & 3 & 0 & 2105 & 0 & 0 & 0 & 0 & 2 & 0 & 0 & 1 & 4 & 2 & 1 & 1 & 0 & 7 \\
& hun & 0 & 2 & 0 & 0 & 0 & 2 & 0 & 4 & 0 & 0 & 0 & 490 & 0 & 0 & 0 & 0 & 0 & 0 & 0 & 0 & 0 & 0 & 0 & 0 & 6 \\
& kor & 0 & 0 & 3 & 0 & 0 & 0 & 0 & 819 & 0 & 0 & 0 & 0 & 106 & 0 & 0 & 0 & 0 & 0 & 0 & 0 & 0 & 1 & 0 & 0 & 2 \\
& msa & 0 & 0 & 0 & 0 & 0 & 0 & 0 & 2 & 0 & 0 & 0 & 0 & 0 & 107 & 0 & 0 & 0 & 0 & 0 & 0 & 0 & 0 & 3 & 0 & 0 \\
& mya & 0 & 0 & 0 & 0 & 0 & 0 & 0 & 128 & 0 & 0 & 0 & 0 & 0 & 0 & 460 & 1 & 0 & 0 & 0 & 0 & 0 & 0 & 0 & 0 & 0 \\
& nld & 0 & 0 & 0 & 0 & 0 & 1 & 0 & 249 & 1 & 0 & 0 & 0 & 0 & 0 & 0 & 1403 & 0 & 0 & 2 & 0 & 0 & 0 & 0 & 0 & 1 \\
& pol & 0 & 0 & 0 & 0 & 0 & 0 & 0 & 122 & 0 & 0 & 0 & 0 & 0 & 0 & 0 & 0 & 102 & 0 & 0 & 0 & 0 & 0 & 0 & 1 & 0 \\
& por & 0 & 0 & 0 & 0 & 0 & 0 & 0 & 3 & 1 & 0 & 0 & 1 & 0 & 0 & 0 & 0 & 0 & 270 & 0 & 0 & 0 & 0 & 0 & 0 & 0 \\
& ron & 0 & 0 & 0 & 0 & 0 & 0 & 2 & 7 & 0 & 0 & 1 & 2 & 0 & 0 & 0 & 0 & 0 & 0 & 459 & 0 & 0 & 0 & 1 & 0 & 1 \\
& sin & 1 & 0 & 1 & 0 & 0 & 0 & 1 & 16 & 32 & 0 & 3 & 1 & 2 & 0 & 0 & 2 & 0 & 0 & 1 & 3739 & 0 & 0 & 1 & 0 & 4 \\
& slk & 0 & 0 & 0 & 0 & 0 & 0 & 1 & 4 & 0 & 0 & 0 & 0 & 0 & 0 & 0 & 0 & 0 & 0 & 0 & 2 & 226 & 0 & 0 & 0 & 1 \\
& spa & 0 & 0 & 0 & 0 & 0 & 0 & 0 & 456 & 0 & 0 & 0 & 0 & 0 & 0 & 0 & 1 & 0 & 0 & 0 & 0 & 0 & 57 & 0 & 0 & 3 \\
& tgl & 0 & 0 & 0 & 0 & 3 & 0 & 0 & 3 & 2 & 0 & 1 & 3 & 0 & 0 & 0 & 0 & 0 & 0 & 0 & 0 & 0 & 0 & 332 & 0 & 0 \\
& tha & 0 & 0 & 0 & 0 & 0 & 0 & 0 & 202 & 0 & 0 & 0 & 0 & 0 & 0 & 0 & 0 & 0 & 0 & 0 & 0 & 0 & 0 & 0 & 574 & 0 \\
& urd & 0 & 0 & 1 & 0 & 0 & 0 & 0 & 328 & 0 & 0 & 1 & 0 & 0 & 0 & 0 & 0 & 0 & 0 & 0 & 0 & 0 & 0 & 0 & 0 & 0 \\
& zho & 0 & 0 & 0 & 0 & 0 & 0 & 0 & 509 & 0 & 0 & 0 & 0 & 0 & 0 & 0 & 17 & 0 & 0 & 0 & 0 & 0 & 0 & 0 & 3 & 15 \\
& Other & 5 & 1 & 0 & 0 & 2 & 5 & 3 & 216 & 20 & 1 & 8 & 3 & 1 & 0 & 1 & 7 & 0 & 3 & 0 & 5 & 1 & 0 & 2 & 2 & 17 \\
\bottomrule
\end{tabular}
\caption{Number of fact-check-to-post pairs for different language combinations. Note that one post can have more than one language assigned.}\label{tab:language_matrix}
\end{table*}

\begin{figure}[t]
\centering
\includegraphics[width=7.5cm]{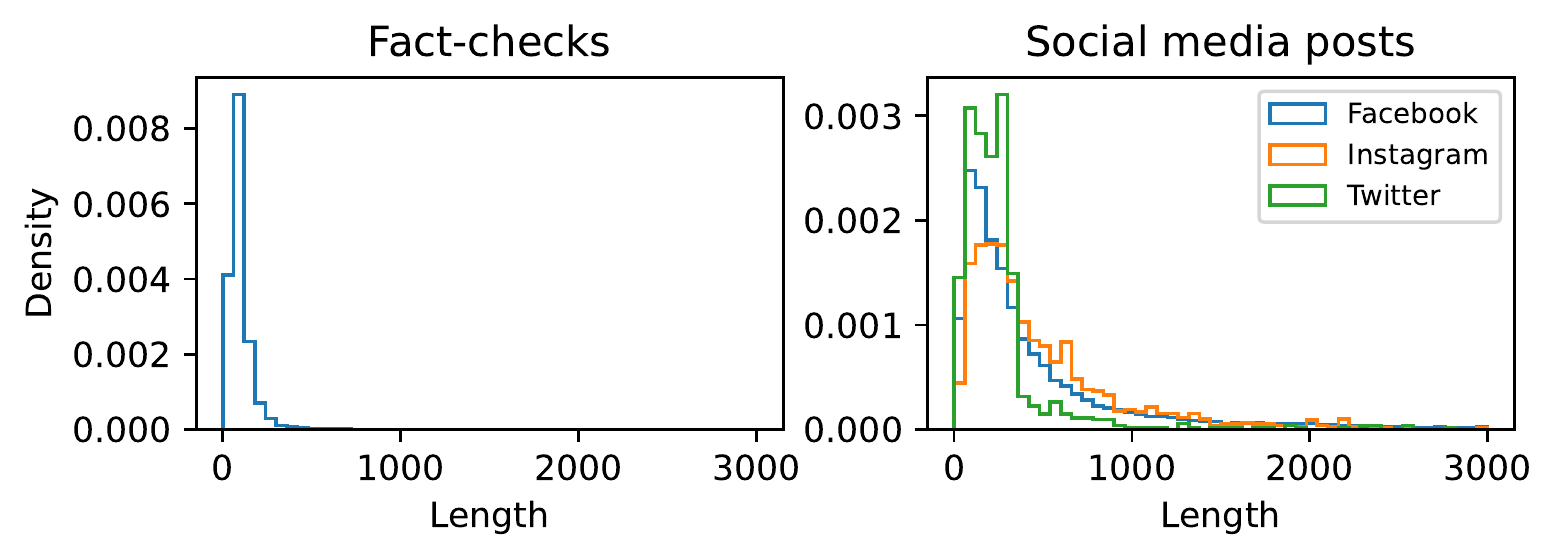}
\caption{Density plots for the character lengths of the fact-checked claims and the social media posts in our dataset.}\label{fig:length_density}
\end{figure}

Figure~\ref{fig:length_density} show the density of lengths for both the fact-checked claims and the posts. Both have long tail distributions, but the claims are in general much shorter. 99\% of claims are shorter than 379 characters. For social media posts, it is 4129 characters.

\section{Hyperparameters}\label{sec:hparams_search}

\subsection{BM25}

We use the default PyTerrier values for BM25 algorithm: $k_1 = 1.2, b = 0.75$. Our preliminary results show that the results are not very sensitive towards these two hyperparameters, probably because of the relatively short length of the documents that we retrieve. Most claims in our dataset have only one sentence.

\subsection{Supervised Training}

Table~\ref{tab:hparams_search} show the range of hyperparmeters in our hyperparameter search done for supervised training in Section~\ref{sec:supervised}, as well as the best performing hyperparameters.
\begin{table*}
    \centering
    \small
    \begin{tabular}{l|p{3cm}|r|r}
    \textbf{Hyperparameter} & \textbf{Range} & \textbf{GTR-T5-Large} & \textbf{MPNet-Base-Multilingual} \\
    \toprule
    Loss & contrastive cosine online-contrastive & online-contrastive & online-contrastive \\
    Learning rate & $[1{e-}3,1{e-}7]$ & $1{e-}5$ & $5{e-}6$ \\
    Learning rate schedule & cosine linear & cosine &  cosine \\
    Warmup steps & $[100, 3200]$ & 800 & 1600 \\
    Weight decay rate & $[1{e-}7, 1{e-}4]$ & $1{e-}5$ & $8{e-}5$ \\
    Ratio of positive to negative samples & $[10, 50\%]$ & 10\% & 30\% \\
    Margin & $[0.1, 0.5]$ & 0.5 & 0.4 \\
    Batch size & Maximum possible & 2 & 8 \\
    \bottomrule
    \end{tabular}
    \caption{Range of hyperparameters used in our supervised hyperparameter search and the hyperparameters of our most successful models. The ranges adjusted during the experimentation according to the preliminary results.}
    \label{tab:hparams_search}
\end{table*}

\section{Additional Results}

\subsection{Additional Metrics}

Table~\ref{tab:metrics} show additional IR metrics calculated for the experiments done in Section~\ref{sec:main}. There is a strong correlation between all these metrics, as shown in Table~\ref{tab:metrics_pearson}. This is caused by the fact that most posts have only one fact-check assigned and the calculations for such cases are very similar for different metrics. We ultimately decided to use S@10 as our main evaluation metric in this work as we find it to be the most interpretable measure (\textit{for how many pairs the expected fact-checked claim ended up in the top 10}).

\begin{table*}
    \centering
    \setlength{\tabcolsep}{3pt} 
    \small
\begin{tabular}{llrrrrrrrrrr}
\toprule
& & \multicolumn{2}{c}{S@10} & \multicolumn{2}{c}{MRR} & \multicolumn{2}{c}{MAP} & \multicolumn{2}{c}{NDCG} & \multicolumn{2}{c}{MAP@10} \\
Method & Ver. & Mono & Cross & Mono & Cross & Mono & Cross & Mono & Cross & Mono & Cross \\
\midrule
\textbf{BM25} \\
BM25 & En & 0.61 & 0.22 & 0.78 & 0.39 & 0.61 & 0.22 & 0.68 & 0.33 & 0.82 & 0.40 \\
BM25 & Og & 0.48 & 0.04 & 0.62 & 0.06 & 0.47 & 0.04 & 0.56 & 0.11 & 0.65 & 0.07 \\ \midrule
\textbf{English TEMs} \\
DistilRoBERTa & En & 0.59 & 0.24 & 0.76 & 0.43 & 0.59 & 0.24 & 0.67 & 0.36 & 0.80 & 0.44 \\
GTR-T5-Base & En & 0.65 & 0.30 & 0.81 & 0.51 & 0.64 & 0.30 & 0.71 & 0.41 & 0.85 & 0.53 \\
GTR-T5-Large & En & \textbf{0.67} & 0.33 & \textbf{0.83} & \textbf{0.56} & \textbf{0.67} & \textbf{0.33} & 0.73 & \textbf{0.45} & \textbf{0.88} & \textbf{0.58} \\
GTR-T5-XL & En & \textbf{0.67} & \textbf{0.34} & \textbf{0.83} & \textbf{0.56} & \textbf{0.67} & \textbf{0.33} & \textbf{0.74} & \textbf{0.45} & 0.87 & \textbf{0.58} \\
MPNet-Base & En & 0.61 & 0.27 & 0.78 & 0.47 & 0.60 & 0.27 & 0.68 & 0.38 & 0.82 & 0.48 \\
MSMARCO-BERT-Base & En & 0.62 & 0.26 & 0.78 & 0.46 & 0.61 & 0.26 & 0.69 & 0.37 & 0.82 & 0.48 \\
MiniLM-L12 & En & 0.64 & 0.29 & 0.80 & 0.48 & 0.63 & 0.29 & 0.71 & 0.40 & 0.84 & 0.50 \\
MultiQA-MPNet-Base & En & 0.64 & 0.29 & 0.80 & 0.50 & 0.63 & 0.29 & 0.71 & 0.41 & 0.84 & 0.52 \\
SGPT-125M & En & 0.47 & 0.14 & 0.63 & 0.25 & 0.47 & 0.14 & 0.56 & 0.25 & 0.66 & 0.26 \\
SGPT-2.7B & En & 0.60 & 0.29 & 0.77 & 0.50 & 0.60 & 0.29 & 0.68 & 0.40 & 0.81 & 0.52 \\
Sentence-T5-Base & En & 0.57 & 0.21 & 0.73 & 0.37 & 0.56 & 0.21 & 0.65 & 0.33 & 0.77 & 0.38 \\
Sentence-T5-Large & En & 0.58 & 0.24 & 0.75 & 0.41 & 0.58 & 0.23 & 0.66 & 0.35 & 0.78 & 0.43 \\
Sentence-T5-XL & En & 0.61 & 0.27 & 0.78 & 0.47 & 0.60 & 0.26 & 0.68 & 0.38 & 0.82 & 0.48 \\ \midrule
\textbf{Multilingual TEMs} \\
DistilUSE-Base-Multilingual & En & 0.58 & 0.22 & 0.74 & 0.40 & 0.58 & 0.22 & 0.66 & 0.34 & 0.78 & 0.41 \\
 & Og & 0.50 & 0.10 & 0.66 & 0.20 & 0.49 & 0.10 & 0.59 & 0.21 & 0.69 & 0.21 \\
LaBSE & En & 0.47 & 0.13 & 0.63 & 0.22 & 0.46 & 0.13 & 0.56 & 0.23 & 0.66 & 0.23 \\
 & Og & 0.53 & 0.12 & 0.69 & 0.22 & 0.53 & 0.12 & 0.61 & 0.22 & 0.72 & 0.23 \\
MPNet-Base-Multilingual & En & \textbf{0.60} & \textbf{0.24} & \textbf{0.75} & \textbf{0.41} & \textbf{0.59} & \textbf{0.23} & \textbf{0.67} & \textbf{0.35} & \textbf{0.79} & \textbf{0.42} \\
 & Og & 0.53 & 0.11 & 0.70 & 0.21 & 0.53 & 0.11 & 0.61 & 0.22 & 0.73 & 0.22 \\
MiniLM-L2-Multilingual & En & 0.58 & 0.22 & 0.74 & 0.38 & 0.57 & 0.22 & 0.66 & 0.34 & 0.77 & 0.40 \\
 & Og & 0.48 & 0.08 & 0.63 & 0.15 & 0.47 & 0.08 & 0.57 & 0.18 & 0.67 & 0.16 \\
XLM-R & En & 0.55 & 0.19 & 0.72 & 0.33 & 0.55 & 0.19 & 0.63 & 0.30 & 0.76 & 0.34 \\
 & Og & 0.50 & 0.08 & 0.66 & 0.15 & 0.50 & 0.08 & 0.59 & 0.18 & 0.70 & 0.16 \\
\bottomrule
\end{tabular}
\caption{The results for different ranking methods. This table shows the same experiment as Table~\ref{tab:main}, but also calculates additional information retrieval metrics: MRR, MAP, NDCG, MAP@10.}\label{tab:metrics}
\end{table*}

\begin{table}
\centering
\tiny
\begin{tabular}{lrrrrr}
\toprule
 & S@10 & MRR & MAP & NDCG & MAP@10 \\
\midrule
S@10 & 1.000 & 0.986 & 0.986 & 0.993 & 1.000 \\
MRR & 0.986 & 1.000 & 1.000 & 0.998 & 0.988 \\
MAP & 0.986 & 1.000 & 1.000 & 0.998 & 0.988 \\
NDCG & 0.993 & 0.998 & 0.998 & 1.000 & 0.995 \\
MAP@10 & 1.000 & 0.988 & 0.988 & 0.995 & 1.000 \\
\bottomrule
\end{tabular}
\caption{Pearson correlation coefficient between different metrics as calculated in Table~\ref{tab:metrics}. Both monolingual and crosslingual scores are taken into consideration.}\label{tab:metrics_pearson}

\end{table}

\subsection{Detailed Per-language Results}

Table~\ref{tab:language_results} shows additional language-specific results for all the methods from Section~\ref{sec:unsupervised} including confidence intervals.

\begin{table*}
\centering
\tiny
    \setlength{\tabcolsep}{3pt} 

\begin{tabular}{llllllllllll}
\toprule
{} & Ver. &          spa &          eng &          por &          fra &          msa &          deu &          ara &          tha &          hbs &          kor \\
\midrule
BM25                        &  En &  0.79 ± 0.01 &  0.62 ± 0.01 &  0.78 ± 0.01 &  0.77 ± 0.02 &  0.74 ± 0.02 &  0.70 ± 0.03 &  0.82 ± 0.03 &  0.89 ± 0.03 &  0.66 ± 0.04 &  0.81 ± 0.04 \\
BM25                        &  Og &  0.76 ± 0.01 &  0.60 ± 0.01 &  0.79 ± 0.01 &  0.71 ± 0.02 &  0.73 ± 0.02 &  0.56 ± 0.03 &  0.75 ± 0.03 &  0.26 ± 0.04 &  0.47 ± 0.04 &  0.58 ± 0.04 \\
DistilRoBERTa               &  En &  0.72 ± 0.01 &  0.64 ± 0.01 &  0.64 ± 0.02 &  0.79 ± 0.02 &  0.75 ± 0.02 &  0.58 ± 0.03 &  0.79 ± 0.03 &  0.89 ± 0.03 &  0.65 ± 0.04 &  0.82 ± 0.04 \\
GTR-T5-Base                 &  En &  0.80 ± 0.01 &  0.72 ± 0.01 &  0.75 ± 0.01 &  0.84 ± 0.02 &  0.80 ± 0.02 &  0.66 ± 0.03 &  0.85 ± 0.02 &  0.90 ± 0.02 &  0.71 ± 0.04 &  0.85 ± 0.03 \\
GTR-T5-Large                &  En &  0.84 ± 0.01 &  0.77 ± 0.01 &  0.80 ± 0.01 &  0.86 ± 0.01 &  0.82 ± 0.02 &  0.69 ± 0.03 &  0.86 ± 0.02 &  0.90 ± 0.02 &  0.74 ± 0.04 &  0.86 ± 0.03 \\
GTR-T5-XL                   &  En &  0.83 ± 0.01 &  0.77 ± 0.01 &  0.80 ± 0.01 &  0.86 ± 0.01 &  0.83 ± 0.02 &  0.70 ± 0.03 &  0.87 ± 0.02 &  0.91 ± 0.02 &  0.73 ± 0.04 &  0.86 ± 0.03 \\
MPNet-Base                  &  En &  0.75 ± 0.01 &  0.68 ± 0.01 &  0.67 ± 0.02 &  0.81 ± 0.02 &  0.79 ± 0.02 &  0.57 ± 0.03 &  0.80 ± 0.03 &  0.88 ± 0.03 &  0.70 ± 0.04 &  0.80 ± 0.04 \\
MSMARCO-BERT-Base           &  En &  0.78 ± 0.01 &  0.63 ± 0.01 &  0.75 ± 0.01 &  0.80 ± 0.02 &  0.79 ± 0.02 &  0.59 ± 0.03 &  0.82 ± 0.03 &  0.87 ± 0.03 &  0.66 ± 0.04 &  0.84 ± 0.03 \\
MiniLM-L12                  &  En &  0.78 ± 0.01 &  0.70 ± 0.01 &  0.71 ± 0.01 &  0.82 ± 0.02 &  0.78 ± 0.02 &  0.64 ± 0.03 &  0.84 ± 0.02 &  0.89 ± 0.03 &  0.72 ± 0.04 &  0.83 ± 0.03 \\
MultiQA-MPNet-Base          &  En &  0.79 ± 0.01 &  0.70 ± 0.01 &  0.71 ± 0.01 &  0.84 ± 0.02 &  0.81 ± 0.02 &  0.64 ± 0.03 &  0.82 ± 0.03 &  0.90 ± 0.02 &  0.71 ± 0.04 &  0.84 ± 0.03 \\
SGPT-125M                   &  En &  0.54 ± 0.01 &  0.40 ± 0.01 &  0.52 ± 0.02 &  0.60 ± 0.02 &  0.55 ± 0.03 &  0.45 ± 0.03 &  0.70 ± 0.03 &  0.82 ± 0.03 &  0.52 ± 0.04 &  0.71 ± 0.04 \\
SGPT-2.7B                   &  En &  0.77 ± 0.01 &  0.65 ± 0.01 &  0.72 ± 0.01 &  0.79 ± 0.02 &  0.79 ± 0.02 &  0.60 ± 0.03 &  0.82 ± 0.03 &  0.91 ± 0.02 &  0.70 ± 0.04 &  0.83 ± 0.03 \\
Sentence-T5-Base            &  En &  0.71 ± 0.01 &  0.60 ± 0.01 &  0.63 ± 0.02 &  0.73 ± 0.02 &  0.77 ± 0.02 &  0.43 ± 0.03 &  0.76 ± 0.03 &  0.89 ± 0.03 &  0.59 ± 0.04 &  0.83 ± 0.03 \\
Sentence-T5-Large           &  En &  0.74 ± 0.01 &  0.64 ± 0.01 &  0.65 ± 0.02 &  0.76 ± 0.02 &  0.76 ± 0.02 &  0.47 ± 0.03 &  0.80 ± 0.03 &  0.89 ± 0.02 &  0.59 ± 0.04 &  0.81 ± 0.04 \\
Sentence-T5-XL              &  En &  0.77 ± 0.01 &  0.68 ± 0.01 &  0.70 ± 0.01 &  0.80 ± 0.02 &  0.79 ± 0.02 &  0.53 ± 0.03 &  0.82 ± 0.03 &  0.91 ± 0.02 &  0.68 ± 0.04 &  0.81 ± 0.04 \\
DistilUSE-Base-Multilingual &  En &  0.69 ± 0.01 &  0.57 ± 0.01 &  0.63 ± 0.02 &  0.75 ± 0.02 &  0.74 ± 0.02 &  0.53 ± 0.03 &  0.80 ± 0.03 &  0.88 ± 0.03 &  0.65 ± 0.04 &  0.79 ± 0.04 \\
DistilUSE-Base-Multilingual &  Og &  0.64 ± 0.01 &  0.56 ± 0.01 &  0.58 ± 0.02 &  0.69 ± 0.02 &  0.60 ± 0.03 &  0.50 ± 0.03 &  0.74 ± 0.03 &  0.72 ± 0.04 &  0.57 ± 0.04 &  0.74 ± 0.04 \\
LaBSE                       &  En &  0.56 ± 0.01 &  0.44 ± 0.01 &  0.53 ± 0.02 &  0.60 ± 0.02 &  0.59 ± 0.03 &  0.34 ± 0.03 &  0.68 ± 0.03 &  0.82 ± 0.03 &  0.44 ± 0.04 &  0.72 ± 0.04 \\
LaBSE                       &  Og &  0.64 ± 0.01 &  0.44 ± 0.01 &  0.66 ± 0.02 &  0.72 ± 0.02 &  0.67 ± 0.02 &  0.48 ± 0.03 &  0.77 ± 0.03 &  0.79 ± 0.03 &  0.57 ± 0.04 &  0.77 ± 0.04 \\
MPNet-Base-Multilingual     &  En &  0.72 ± 0.01 &  0.62 ± 0.01 &  0.64 ± 0.02 &  0.79 ± 0.02 &  0.73 ± 0.02 &  0.58 ± 0.03 &  0.83 ± 0.03 &  0.89 ± 0.03 &  0.64 ± 0.04 &  0.80 ± 0.04 \\
MPNet-Base-Multilingual     &  Og &  0.64 ± 0.01 &  0.61 ± 0.01 &  0.57 ± 0.02 &  0.73 ± 0.02 &  0.61 ± 0.03 &  0.53 ± 0.03 &  0.70 ± 0.03 &  0.86 ± 0.03 &  0.56 ± 0.04 &  0.71 ± 0.04 \\
MiniLM-L2-Multilingual      &  En &  0.69 ± 0.01 &  0.59 ± 0.01 &  0.60 ± 0.02 &  0.75 ± 0.02 &  0.71 ± 0.02 &  0.54 ± 0.03 &  0.82 ± 0.03 &  0.86 ± 0.03 &  0.65 ± 0.04 &  0.79 ± 0.04 \\
MiniLM-L2-Multilingual      &  Og &  0.57 ± 0.01 &  0.57 ± 0.01 &  0.51 ± 0.02 &  0.66 ± 0.02 &  0.54 ± 0.03 &  0.49 ± 0.03 &  0.49 ± 0.03 &  0.82 ± 0.03 &  0.55 ± 0.04 &  0.61 ± 0.04 \\
XLM-R                       &  En &  0.64 ± 0.01 &  0.53 ± 0.01 &  0.60 ± 0.02 &  0.72 ± 0.02 &  0.69 ± 0.02 &  0.56 ± 0.03 &  0.81 ± 0.03 &  0.86 ± 0.03 &  0.58 ± 0.04 &  0.79 ± 0.04 \\
XLM-R                       &  Og &  0.58 ± 0.01 &  0.52 ± 0.01 &  0.55 ± 0.02 &  0.70 ± 0.02 &  0.57 ± 0.03 &  0.54 ± 0.03 &  0.51 ± 0.03 &  0.82 ± 0.03 &  0.54 ± 0.04 &  0.63 ± 0.04 \\
\midrule
{} & Ver. &          pol &          slk &          nld &          ron &          ell &          ces &          bul &          hun &          hin &          mya \\
\midrule
BM25                        &  En &  0.69 ± 0.04 &  0.83 ± 0.04 &  0.73 ± 0.05 &  0.84 ± 0.05 &  0.76 ± 0.06 &  0.85 ± 0.05 &  0.81 ± 0.07 &  0.81 ± 0.07 &  0.73 ± 0.08 &  0.78 ± 0.08 \\
BM25                        &  Og &  0.60 ± 0.04 &  0.69 ± 0.05 &  0.57 ± 0.06 &  0.71 ± 0.06 &  0.66 ± 0.06 &  0.69 ± 0.06 &  0.70 ± 0.08 &  0.51 ± 0.09 &  0.61 ± 0.09 &  0.32 ± 0.09 \\
DistilRoBERTa               &  En &  0.65 ± 0.04 &  0.85 ± 0.04 &  0.72 ± 0.05 &  0.86 ± 0.05 &  0.73 ± 0.06 &  0.88 ± 0.04 &  0.86 ± 0.06 &  0.82 ± 0.07 &  0.65 ± 0.09 &  0.77 ± 0.08 \\
GTR-T5-Base                 &  En &  0.68 ± 0.04 &  0.87 ± 0.04 &  0.74 ± 0.05 &  0.88 ± 0.04 &  0.78 ± 0.05 &  0.90 ± 0.04 &  0.86 ± 0.06 &  0.87 ± 0.06 &  0.76 ± 0.08 &  0.85 ± 0.07 \\
GTR-T5-Large                &  En &  0.74 ± 0.04 &  0.87 ± 0.04 &  0.78 ± 0.05 &  0.88 ± 0.04 &  0.83 ± 0.05 &  0.88 ± 0.04 &  0.86 ± 0.06 &  0.89 ± 0.06 &  0.79 ± 0.08 &  0.88 ± 0.06 \\
GTR-T5-XL                   &  En &  0.70 ± 0.04 &  0.88 ± 0.03 &  0.80 ± 0.05 &  0.87 ± 0.04 &  0.85 ± 0.05 &  0.88 ± 0.04 &  0.87 ± 0.06 &  0.86 ± 0.07 &  0.80 ± 0.08 &  0.84 ± 0.07 \\
MPNet-Base                  &  En &  0.67 ± 0.04 &  0.84 ± 0.04 &  0.74 ± 0.05 &  0.86 ± 0.04 &  0.77 ± 0.05 &  0.89 ± 0.04 &  0.87 ± 0.06 &  0.87 ± 0.06 &  0.72 ± 0.08 &  0.80 ± 0.08 \\
MSMARCO-BERT-Base           &  En &  0.64 ± 0.04 &  0.85 ± 0.04 &  0.66 ± 0.06 &  0.83 ± 0.05 &  0.77 ± 0.06 &  0.86 ± 0.05 &  0.87 ± 0.06 &  0.85 ± 0.07 &  0.80 ± 0.08 &  0.82 ± 0.07 \\
MiniLM-L12                  &  En &  0.72 ± 0.04 &  0.86 ± 0.04 &  0.73 ± 0.05 &  0.86 ± 0.04 &  0.80 ± 0.05 &  0.86 ± 0.05 &  0.90 ± 0.05 &  0.86 ± 0.07 &  0.77 ± 0.08 &  0.80 ± 0.08 \\
MultiQA-MPNet-Base          &  En &  0.70 ± 0.04 &  0.86 ± 0.04 &  0.75 ± 0.05 &  0.87 ± 0.04 &  0.79 ± 0.05 &  0.87 ± 0.04 &  0.89 ± 0.06 &  0.87 ± 0.06 &  0.73 ± 0.08 &  0.84 ± 0.07 \\
SGPT-125M                   &  En &  0.54 ± 0.05 &  0.71 ± 0.05 &  0.60 ± 0.06 &  0.75 ± 0.06 &  0.54 ± 0.06 &  0.73 ± 0.06 &  0.76 ± 0.08 &  0.78 ± 0.08 &  0.50 ± 0.09 &  0.75 ± 0.08 \\
SGPT-2.7B                   &  En &  0.65 ± 0.04 &  0.83 ± 0.04 &  0.70 ± 0.05 &  0.83 ± 0.05 &  0.76 ± 0.06 &  0.85 ± 0.05 &  0.87 ± 0.06 &  0.83 ± 0.07 &  0.69 ± 0.09 &  0.82 ± 0.07 \\
Sentence-T5-Base            &  En &  0.57 ± 0.05 &  0.84 ± 0.04 &  0.62 ± 0.06 &  0.87 ± 0.04 &  0.76 ± 0.06 &  0.82 ± 0.05 &  0.87 ± 0.06 &  0.84 ± 0.07 &  0.70 ± 0.09 &  0.76 ± 0.08 \\
Sentence-T5-Large           &  En &  0.60 ± 0.04 &  0.86 ± 0.04 &  0.62 ± 0.06 &  0.86 ± 0.04 &  0.77 ± 0.05 &  0.86 ± 0.05 &  0.85 ± 0.07 &  0.82 ± 0.07 &  0.68 ± 0.09 &  0.80 ± 0.08 \\
Sentence-T5-XL              &  En &  0.66 ± 0.04 &  0.87 ± 0.04 &  0.70 ± 0.05 &  0.87 ± 0.04 &  0.81 ± 0.05 &  0.86 ± 0.05 &  0.87 ± 0.06 &  0.83 ± 0.07 &  0.73 ± 0.08 &  0.80 ± 0.08 \\
DistilUSE-Base-Multilingual &  En &  0.68 ± 0.04 &  0.82 ± 0.04 &  0.70 ± 0.05 &  0.84 ± 0.05 &  0.76 ± 0.06 &  0.78 ± 0.06 &  0.89 ± 0.06 &  0.83 ± 0.07 &  0.63 ± 0.09 &  0.80 ± 0.08 \\
DistilUSE-Base-Multilingual &  Og &  0.60 ± 0.04 &  0.76 ± 0.05 &  0.61 ± 0.06 &  0.80 ± 0.05 &  0.60 ± 0.06 &  0.71 ± 0.06 &  0.81 ± 0.07 &  0.78 ± 0.08 &  0.53 ± 0.09 &  0.62 ± 0.09 \\
LaBSE                       &  En &  0.45 ± 0.05 &  0.73 ± 0.05 &  0.51 ± 0.06 &  0.77 ± 0.05 &  0.67 ± 0.06 &  0.80 ± 0.05 &  0.78 ± 0.08 &  0.83 ± 0.07 &  0.55 ± 0.09 &  0.76 ± 0.08 \\
LaBSE                       &  Og &  0.57 ± 0.05 &  0.74 ± 0.05 &  0.61 ± 0.06 &  0.78 ± 0.05 &  0.70 ± 0.06 &  0.81 ± 0.05 &  0.84 ± 0.07 &  0.82 ± 0.07 &  0.56 ± 0.09 &  0.77 ± 0.08 \\
MPNet-Base-Multilingual     &  En &  0.64 ± 0.04 &  0.84 ± 0.04 &  0.71 ± 0.05 &  0.86 ± 0.05 &  0.75 ± 0.06 &  0.82 ± 0.05 &  0.85 ± 0.07 &  0.82 ± 0.07 &  0.70 ± 0.09 &  0.77 ± 0.08 \\
MPNet-Base-Multilingual     &  Og &  0.60 ± 0.04 &  0.79 ± 0.04 &  0.66 ± 0.06 &  0.84 ± 0.05 &  0.63 ± 0.06 &  0.78 ± 0.06 &  0.81 ± 0.07 &  0.83 ± 0.07 &  0.63 ± 0.09 &  0.75 ± 0.08 \\
MiniLM-L2-Multilingual      &  En &  0.66 ± 0.04 &  0.83 ± 0.04 &  0.74 ± 0.05 &  0.81 ± 0.05 &  0.75 ± 0.06 &  0.79 ± 0.05 &  0.87 ± 0.06 &  0.81 ± 0.07 &  0.62 ± 0.09 &  0.79 ± 0.08 \\
MiniLM-L2-Multilingual      &  Og &  0.61 ± 0.04 &  0.77 ± 0.04 &  0.64 ± 0.06 &  0.79 ± 0.05 &  0.58 ± 0.06 &  0.75 ± 0.06 &  0.83 ± 0.07 &  0.79 ± 0.08 &  0.49 ± 0.09 &  0.58 ± 0.09 \\
XLM-R                       &  En &  0.59 ± 0.04 &  0.82 ± 0.04 &  0.64 ± 0.06 &  0.84 ± 0.05 &  0.71 ± 0.06 &  0.77 ± 0.06 &  0.89 ± 0.06 &  0.78 ± 0.08 &  0.71 ± 0.09 &  0.81 ± 0.07 \\
XLM-R                       &  Og &  0.59 ± 0.04 &  0.80 ± 0.04 &  0.65 ± 0.06 &  0.79 ± 0.05 &  0.65 ± 0.06 &  0.78 ± 0.06 &  0.82 ± 0.07 &  0.80 ± 0.07 &  0.62 ± 0.09 &  0.75 ± 0.08 \\
\midrule
{} & Ver. &      eng-hin &      eng-zho &      eng-sin &      eng-urd &      eng-tgl &      eng-msa &      eng-tha &      eng-kor &      eng-mya &        Other \\
\midrule
BM25                        &  En &  0.35 ± 0.03 &  0.46 ± 0.04 &  0.32 ± 0.04 &  0.35 ± 0.05 &  0.37 ± 0.06 &  0.45 ± 0.06 &  0.42 ± 0.07 &  0.42 ± 0.08 &  0.38 ± 0.08 &  0.36 ± 0.03 \\
BM25                        &  Og &  0.04 ± 0.01 &  0.02 ± 0.01 &  0.03 ± 0.01 &  0.01 ± 0.01 &  0.11 ± 0.04 &  0.08 ± 0.03 &  0.09 ± 0.04 &  0.03 ± 0.03 &  0.10 ± 0.05 &  0.18 ± 0.02 \\
DistilRoBERTa               &  En &  0.33 ± 0.03 &  0.50 ± 0.04 &  0.34 ± 0.04 &  0.38 ± 0.05 &  0.37 ± 0.06 &  0.53 ± 0.06 &  0.56 ± 0.07 &  0.48 ± 0.09 &  0.43 ± 0.09 &  0.34 ± 0.03 \\
GTR-T5-Base                 &  En &  0.44 ± 0.03 &  0.61 ± 0.04 &  0.41 ± 0.05 &  0.47 ± 0.05 &  0.46 ± 0.06 &  0.62 ± 0.06 &  0.50 ± 0.07 &  0.61 ± 0.08 &  0.52 ± 0.09 &  0.43 ± 0.03 \\
GTR-T5-Large                &  En &  0.51 ± 0.03 &  0.67 ± 0.04 &  0.48 ± 0.05 &  0.51 ± 0.05 &  0.56 ± 0.06 &  0.63 ± 0.06 &  0.59 ± 0.07 &  0.62 ± 0.08 &  0.52 ± 0.09 &  0.46 ± 0.03 \\
GTR-T5-XL                   &  En &  0.51 ± 0.03 &  0.66 ± 0.04 &  0.45 ± 0.05 &  0.56 ± 0.05 &  0.53 ± 0.06 &  0.66 ± 0.06 &  0.60 ± 0.07 &  0.60 ± 0.08 &  0.53 ± 0.09 &  0.47 ± 0.03 \\
MPNet-Base                  &  En &  0.41 ± 0.03 &  0.52 ± 0.04 &  0.36 ± 0.04 &  0.42 ± 0.05 &  0.44 ± 0.06 &  0.60 ± 0.06 &  0.55 ± 0.07 &  0.54 ± 0.09 &  0.42 ± 0.09 &  0.40 ± 0.03 \\
MSMARCO-BERT-Base           &  En &  0.42 ± 0.03 &  0.52 ± 0.04 &  0.38 ± 0.04 &  0.38 ± 0.05 &  0.41 ± 0.06 &  0.55 ± 0.06 &  0.50 ± 0.07 &  0.51 ± 0.09 &  0.50 ± 0.09 &  0.40 ± 0.03 \\
MiniLM-L12                  &  En &  0.45 ± 0.03 &  0.54 ± 0.04 &  0.40 ± 0.04 &  0.46 ± 0.05 &  0.45 ± 0.06 &  0.60 ± 0.06 &  0.55 ± 0.07 &  0.54 ± 0.09 &  0.44 ± 0.09 &  0.41 ± 0.03 \\
MultiQA-MPNet-Base          &  En &  0.45 ± 0.03 &  0.57 ± 0.04 &  0.41 ± 0.04 &  0.52 ± 0.05 &  0.46 ± 0.06 &  0.61 ± 0.06 &  0.56 ± 0.07 &  0.55 ± 0.08 &  0.46 ± 0.09 &  0.42 ± 0.03 \\
SGPT-125M                   &  En &  0.24 ± 0.03 &  0.30 ± 0.04 &  0.24 ± 0.04 &  0.23 ± 0.05 &  0.19 ± 0.05 &  0.29 ± 0.06 &  0.30 ± 0.06 &  0.26 ± 0.08 &  0.24 ± 0.07 &  0.21 ± 0.02 \\
SGPT-2.7B                   &  En &  0.42 ± 0.03 &  0.58 ± 0.04 &  0.40 ± 0.04 &  0.52 ± 0.05 &  0.45 ± 0.06 &  0.59 ± 0.06 &  0.55 ± 0.07 &  0.53 ± 0.09 &  0.52 ± 0.09 &  0.40 ± 0.03 \\
Sentence-T5-Base            &  En &  0.35 ± 0.03 &  0.43 ± 0.04 &  0.29 ± 0.04 &  0.35 ± 0.05 &  0.36 ± 0.06 &  0.52 ± 0.06 &  0.40 ± 0.07 &  0.37 ± 0.08 &  0.32 ± 0.08 &  0.32 ± 0.03 \\
Sentence-T5-Large           &  En &  0.38 ± 0.03 &  0.48 ± 0.04 &  0.34 ± 0.04 &  0.38 ± 0.05 &  0.44 ± 0.06 &  0.59 ± 0.06 &  0.44 ± 0.07 &  0.39 ± 0.08 &  0.37 ± 0.08 &  0.35 ± 0.03 \\
Sentence-T5-XL              &  En &  0.43 ± 0.03 &  0.52 ± 0.04 &  0.38 ± 0.04 &  0.45 ± 0.05 &  0.49 ± 0.06 &  0.63 ± 0.06 &  0.53 ± 0.07 &  0.46 ± 0.09 &  0.38 ± 0.08 &  0.38 ± 0.03 \\
DistilUSE-Base-Multilingual &  En &  0.38 ± 0.03 &  0.49 ± 0.04 &  0.33 ± 0.04 &  0.38 ± 0.05 &  0.28 ± 0.05 &  0.55 ± 0.06 &  0.44 ± 0.07 &  0.46 ± 0.09 &  0.35 ± 0.08 &  0.32 ± 0.03 \\
DistilUSE-Base-Multilingual &  Og &  0.23 ± 0.03 &  0.35 ± 0.04 &  0.02 ± 0.01 &  0.25 ± 0.05 &  0.11 ± 0.04 &  0.38 ± 0.06 &  0.16 ± 0.05 &  0.20 ± 0.07 &  0.10 ± 0.05 &  0.23 ± 0.02 \\
LaBSE                       &  En &  0.33 ± 0.03 &  0.23 ± 0.04 &  0.18 ± 0.04 &  0.30 ± 0.05 &  0.13 ± 0.04 &  0.28 ± 0.06 &  0.17 ± 0.05 &  0.19 ± 0.07 &  0.24 ± 0.07 &  0.23 ± 0.02 \\
LaBSE                       &  Og &  0.24 ± 0.03 &  0.30 ± 0.04 &  0.20 ± 0.04 &  0.22 ± 0.04 &  0.15 ± 0.04 &  0.28 ± 0.06 &  0.13 ± 0.05 &  0.23 ± 0.07 &  0.22 ± 0.07 &  0.26 ± 0.02 \\
MPNet-Base-Multilingual     &  En &  0.34 ± 0.03 &  0.49 ± 0.04 &  0.32 ± 0.04 &  0.39 ± 0.05 &  0.34 ± 0.06 &  0.57 ± 0.06 &  0.50 ± 0.07 &  0.42 ± 0.08 &  0.33 ± 0.08 &  0.37 ± 0.03 \\
MPNet-Base-Multilingual     &  Og &  0.20 ± 0.03 &  0.31 ± 0.04 &  0.08 ± 0.03 &  0.23 ± 0.05 &  0.10 ± 0.04 &  0.35 ± 0.06 &  0.30 ± 0.06 &  0.18 ± 0.07 &  0.16 ± 0.06 &  0.27 ± 0.02 \\
MiniLM-L2-Multilingual      &  En &  0.33 ± 0.03 &  0.43 ± 0.04 &  0.31 ± 0.04 &  0.33 ± 0.05 &  0.30 ± 0.05 &  0.55 ± 0.06 &  0.48 ± 0.07 &  0.39 ± 0.08 &  0.36 ± 0.08 &  0.34 ± 0.03 \\
MiniLM-L2-Multilingual      &  Og &  0.13 ± 0.02 &  0.24 ± 0.04 &  0.02 ± 0.01 &  0.15 ± 0.04 &  0.08 ± 0.03 &  0.29 ± 0.06 &  0.24 ± 0.06 &  0.12 ± 0.06 &  0.08 ± 0.05 &  0.21 ± 0.02 \\
XLM-R                       &  En &  0.31 ± 0.03 &  0.39 ± 0.04 &  0.30 ± 0.04 &  0.36 ± 0.05 &  0.25 ± 0.05 &  0.45 ± 0.06 &  0.36 ± 0.07 &  0.36 ± 0.08 &  0.27 ± 0.08 &  0.29 ± 0.03 \\
XLM-R                       &  Og &  0.20 ± 0.03 &  0.22 ± 0.04 &  0.05 ± 0.02 &  0.20 ± 0.04 &  0.06 ± 0.03 &  0.26 ± 0.05 &  0.16 ± 0.05 &  0.12 ± 0.06 &  0.13 ± 0.06 &  0.20 ± 0.02 \\
\bottomrule
\end{tabular}

\caption{S@10 performance with confidence intervals for individual languages. This table has the same results as Figure~\ref{fig:languages}, but also calculated for additional methods.}\label{tab:language_results}
\end{table*}

\section{Other Ideas}

Here we discuss some additional ideas that were tried and that we decided not to include in the main text for various reasons.

\paragraph{Sliding window embedding.} Figure~\ref{fig:length} shows that the performance for methods decreases for posts with certain length. The decrease is generally starting at around 500 characters. We experimented with using sliding windows with various sizes (both based on the number of characters and the number of sentences) and strides. TEMs then encode only this sliding window and the final vector similarity is calculated as the maximum similarity of any of the windows. We found out that this technique can slightly ($+0.01-0.02$ S@10) improve the results for TEMs.

\paragraph{Using fact-check titles alongside claims.}
We represent fact-checks with the \textit{claim} field obtained from the data in our main text experiments. We also experimented with the \textit{title} field that we were able to obtain for the majority of the fact-checks. We found out that representing the fact-check as a concatenation of a claim and a title improves the results slightly ($+0.00-0.01$ S@10) for BM25 methods.

\paragraph{Topic detection.} We attempted to run a topic detection over our posts to better understand how different methods handle different topics and themes in our data. We experimented with both \textit{original} and \textit{English} versions, with both multilingual and monolingual topic detection models, such as LDA~\citep{blei2003latent} or BERTopic~\citep{bertopic}. Ultimately we were not content with the quality of topic detection, as the models failed to reliably identify even the most frequent topics in our data, such as the COVID-19 pandemic or Russo-Ukrainian war. We believe that this is caused by the short length of the majority of the posts, as well as their relatively noisy nature.

\paragraph{Mixing original and English versions.}
We experimented with representing both fact-checks and posts as a concatenation of both the original language texts and the English translations, so that the multilingual methods can use both sources of information. However, this increased the \textit{same language bias} significantly while the performance decreased significantly across the board.

\section{Examples}\label{app:examples}

This Appendix contains 5 randomly selected fact-check-post pairs from our dataset. We show here all the information present in our dataset for these samples.

\subsection{Example \#1}
\subsubsection{Fact-check}
\textbf{ID:} 104315 \\
\textbf{Published at:} 2021-09-27 \href{https://factcheck.afp.com/http%253A%252F%252Fdoc.afp.com%252F9ND3KL-1#4965c1dd0a82175969c22c4089ed0d0e}{factcheck.afp.com} \\
\\ \textbf{Claim}\\
\textbf{Original text:} \textit{Photo shows meeting of five international intelligence agencies in Delhi}\\
\textbf{Translated text:} \textit{Photo shows meeting of five international intelligence agencies in Delhi}\\
\textbf{Detected languages:} eng: 100.0\% \\
\\ \textbf{Title}\\
\textbf{Original text:} \textit{This photo shows a delegation-level meeting between Indian and Russian national security advisors}\\
\textbf{Translated text:} \textit{This photo shows a delegation-level meeting between Indian and Russian national security advisors}\\
\textbf{Detected languages:} eng: 100.0\% \\

\subsubsection{Social Media Post}
\textbf{ID:} 16806 \\
\textbf{Published at:} Facebook 2021-09-16 \\
\textbf{Verdicts:} Partly false information \\
\\ \textbf{Main text} \\
\textbf{Original text:} \textit{Post Giri IyerNow in Delhi !!India   RAWIsrael   MOSSADAmerica   CIARussia   KGBEngland   MI6First time ever that the top five intelligence agency of the world are sitting together for a high level meeting in Delhi. This is the power of new India}\\
\textbf{Translated text:} \textit{Post Giri IyerNow in Delhi !!India   RAWIsrael   MOSSADAmerica   CIARussia   KGBEngland   MI6First time ever that the top five intelligence agency of the world are sitting together for a high level meeting in Delhi. This is the power of new India}\\
\textbf{Detected languages:} eng: 100.0\% \\
\subsection{Example \#2}
\subsubsection{Fact-check}
\textbf{ID:} 34296 \\
\textbf{Published at:} 2019-10-29 \href{https://factuel.afp.com/non-cette-boisson-base-de-papaye-de-citron-de-racines-de-cocotier-et-de-moringa-bouillis-ne-guerit#f3c5970290e8bacd3e437edff1563f95}{factuel.afp.com} \\
\\ \textbf{Claim}\\
\textbf{Original text:} \textit{COMMENT TRAITER L’HÉPATITE B PAR LES PLANTES}\\
\textbf{Translated text:} \textit{HOW TO TREAT HEPATITIS B WITH HERBS}\\
\textbf{Detected languages:} fra: 100.0\% \\
\\ \textbf{Title}\\
\textbf{Original text:} \textit{Non, cette boisson à base de papaye, de citron, de racines de cocotier et de moringa bouillis, ne guérit pas l'hépatite B}\\
\textbf{Translated text:} \textit{No, this drink made from boiled papaya, lemon, coconut palm roots and moringa does not cure hepatitis B}\\
\textbf{Detected languages:} fra: 100.0\% \\

\subsubsection{Social Media Post}
\textbf{ID:} 11569 \\
\textbf{Published at:} Facebook 2019-07-06 \\
\textbf{Verdicts:} False information \\
\\ \textbf{Main text} \\
\textbf{Original text:} \textit{HÉPATITE B ET Cet très rassurant faite cette expérience et rêvené témoigné !!! L hépatite n'es qu'un vieux souvenirs après !!REMÈDES POUR TRAITER ET ÉRADIQUER L'HÉPATITE B DU CORPSL'hépatite B est une infection virale qui s'attaque au foie.Le virus se transmet par le sang ou lors des rapports sexuels. En effet, les seules sécrétions ou liquides corporels qui permettent de transmettre le virus sont le sang, le sperme,les sécrétions vaginales, la salive et les liquides issus d'une plaieIngrédients Une Papaye non mur Les RACINES de Papayer Femelle Les feuilles fraîches de Papayer femelle racines de Moringa  feuilles fraîche Moringa 4 citrons à couper en deux. racines de cocotierPréparationMettez les tous dans la marmite, les feuilles en dernier position. Ajoutez de l'eau et faites bouillir le mélange.Mode d'emploiBoire 2 à 3 verres par jour.Ajoutez de l'eau à chaque fois et faites bouillir une fois par jour . Suivez le traitement pendant un mois.Faites vous examiné par un médecin et revenez témoigner .Bonne guérison...Aimes ton prochain par le partage de ce messageLa Boutique du Naturopathe Vous soigne de toutes vos maladies à l’aide des plantes naturelles moins chère et plus sure sans effets secondaire.}\\
\textbf{Translated text:} \textit{HEPATITIS B AND Cand very reassuring made this experience and dreamed witnessed !!!Hepatitis is just an old memory afterwards!!REMEDIES TO TREAT AND ERADICATE HEPATITIS B FROM THE BODYHepatitis B is a viral infection that attacks the liver. The virus is transmitted through blood or during sexual intercourse. Indeed, the only secretions or bodily fluids that can transmit the virus are blood, semen, vaginal secretions, saliva and fluids from a wound.Ingredients An unripe Papaya The ROOTS of Female Papaya Fresh female papaya leaves Moringa roots fresh Moringa leaves 4 lemons to be cut in half. coconut rootsPreparationPut them all in the pot, the leaves last. Add water and boil the mixture.ManualDrink 2-3 glasses a day. Add water each time and boil once a day. Follow the treatment for a month. Get examined by a doctor and come back to testify.Good recovery...Love your neighbor by sharing this messageLa Boutique du Naturopathe Treats you to all your illnesses using cheaper and safer natural plants without side effects.}\\
\textbf{Detected languages:} fra: 100.0\% \\
\\ \textbf{OCR transcripts} \\
\textbf{Original text:} \textit{PoymeraseVirus del'hépatite BParticule filamenteuseADNAntigèneHBSParticule sphérique}\\
\textbf{Translated text:} \textit{Polymerasevirushepatitis BFilamentous particleDNAAntigenHBSspherical particle}\\
\textbf{Detected languages:} fra: 72.4\%, lb: 9.2\% \\
\subsection{Example \#3}
\subsubsection{Fact-check}
\textbf{ID:} 93800 \\
\textbf{Published at:} 2021-12-07 \href{https://factual.afp.com/http%253A%252F%252Fdoc.afp.com%252F9U74EQ-1#a02abd86edf541787e02cd30fe4b25ba}{factual.afp.com} \\
\\ \textbf{Claim}\\
\textbf{Original text:} \textit{Nicolás Maduro se fotografió con una camiseta del candidato chileno Gabriel Boric}\\
\textbf{Translated text:} \textit{Nicolás Maduro was photographed with a shirt of the Chilean candidate Gabriel Boric}\\
\textbf{Detected languages:} spa: 100.0\% \\
\\ \textbf{Title}\\
\textbf{Original text:} \textit{El tuit de Maduro con una camiseta del candidato chileno Gabriel Boric es un doble montaje}\\
\textbf{Translated text:} \textit{Maduro's tweet with a t-shirt of the Chilean candidate Gabriel Boric is a double montage}\\
\textbf{Detected languages:} spa: 100.0\% \\

\subsubsection{Social Media Post}
\textbf{ID:} 20617 \\
\textbf{Published at:} Facebook 2021-12-03 \\
\textbf{Verdicts:} Altered photo \\
\\ \textbf{Main text} \\
\textbf{Original text:} \textit{Vamos con esos apoyos Gabrielito}\\
\textbf{Translated text:} \textit{Let's go with those support Gabrielito}\\
\textbf{Detected languages:} spa: 100.0\% \\
\\ \textbf{OCR transcripts} \\
\textbf{Original text:} \textit{Nicolás Maduro@Nicolas MaduroPor la patria grande, nuestro totalapoyo desde Venezuela al compañeroGabriel Boric.Tik Tok \#horicchanta \#horiccorrupto}\\
\textbf{Translated text:} \textit{Nicholas Maduro@NicolasMaduroFor the great country, our totalsupport from Venezuela to the comradeGabriel Boric.tik tok \#horicchanta \#horiccorrupt}\\
\textbf{Detected languages:} spa: 50.4\%, eng: 31.6\%, qu: 10.0\% \\
\subsection{Example \#4}
\subsubsection{Fact-check}
\textbf{ID:} 26926 \\
\textbf{Published at:} 2022-03-16 \href{https://checamos.afp.com/doc.afp.com.326B6NN#f4bb97d6ad037f4f15b275c2c0164e4c}{checamos.afp.com} \\
\\ \textbf{Claim}\\
\textbf{Original text:} \textit{As fronteiras da Ucrânia não foram registradas na ONU e não são reconhecidas internacionalmente}\\
\textbf{Translated text:} \textit{Ukraine's borders have not been registered with the UN and are not internationally recognized}\\
\textbf{Detected languages:} por: 100.0\% \\
\\ \textbf{Title}\\
\textbf{Original text:} \textit{As fronteiras da Ucrânia são reconhecidas e não é preciso que sejam registradas na ONU}\\
\textbf{Translated text:} \textit{Ukraine's borders are recognized and do not need to be registered with the UN}\\
\textbf{Detected languages:} por: 100.0\% \\

\subsubsection{Social Media Post}
\textbf{ID:} 8853 \\
\textbf{Published at:} Facebook 2022-02-25 \\
\textbf{Verdicts:} False information. \\
\\ \textbf{Main text} \\
\textbf{Original text:} \textit{E esta hein..??!!... haverá contraditório..??.....'' O secretário-geral das Nações Unidas afirmou que a Ucrânia não solicita registro de fronteira desde 1991, então o estado da Ucrânia não existe.... E não sabemos disso!!! 04/07/2014 O secretário-geral da ONU, Ban Ki-moon, fez uma declaração impressionante, cuja distribuição na mídia ucraniana e na Internet está proibida.  O conflito entre os dois países foi discutido na sessão do Conselho de Segurança da ONU.  A partir disso, chegou-se à seguinte conclusão: A Ucrânia não registra suas fronteiras desde 25/12/1991.  A ONU não registrou as fronteiras da Ucrânia como um estado soberano. Portanto, pode-se supor que a Rússia não está cometendo nenhuma violação de direitos em relação à Ucrânia. De acordo com o Tratado da CEI, o território da Ucrânia é um distrito administrativo da URSS.  Portanto, ninguém pode ser culpado pelo separatismo e pela mudança forçada das fronteiras da Ucrânia. Sob a lei internacional, o país simplesmente não tem fronteiras oficialmente reconhecidas. Para resolver esse problema, a Ucrânia precisa concluir a demarcação das fronteiras com os países vizinhos e obter o acordo dos países vizinhos, incluindo a Rússia, em sua fronteira comum.  É necessário documentar tudo e assinar tratados com todos os estados vizinhos. A União Europeia prometeu o seu apoio à Ucrânia nesta importante questão e decidiu prestar toda a assistência técnica. Mas a Rússia assinará um tratado de fronteira com a Ucrânia?  Não, claro que não Como a Rússia é a sucessora legal da URSS (isso é confirmado pelas decisões dos tribunais internacionais sobre disputas de propriedade entre a ex-URSS e países estrangeiros), as terras em que a Ucrânia, a Bielorrússia e a Novorossiya estão localizadas pertencem à Rússia, e ninguém tem o  direito de ficar sem o consentimento da Rússia para dispor desta área. Basicamente, agora tudo o que a Rússia precisa fazer é declarar que essa área é russa e que tudo o que acontece nessa área é um assunto interno da Rússia. Qualquer interferência será vista como uma medida contra a Rússia.  Com base nisso, eles podem anular as eleições de 25 de maio de 2014 e fazer o que o povo quiser! De acordo com o Memorando de Budapeste e outros acordos, a Ucrânia não tem fronteiras.  O estado da Ucrânia não existe (e nunca existiu!)..'' Alexandre Panin}\\
\textbf{Translated text:} \textit{And this one huh..??!!...there will be a contradiction..??.....'' The Secretary-General of the United Nations stated that Ukraine has not applied for border registration since 1991, so the state of Ukraine does not exists.... And we don't know that!!! 04/07/2014 The Secretary-General of the UN, Ban Ki-moon, made an impressive statement, whose distribution in the Ukrainian media and on the Internet is prohibited. The conflict between the two countries was discussed at the UN Security Council session. From this, the following conclusion was reached: Ukraine has not registered its borders since 12/25/1991. The UN has not registered Ukraine's borders as a sovereign state. Therefore, it can be assumed that Russia is not committing any rights violations in relation to Ukraine. According to the CIS Treaty, the territory of Ukraine is an administrative district of USSR. Therefore, no one can be blamed for separatism and the forced change of Ukraine's borders. Under international law, the country simply has no officially recognized borders. To solve this problem, Ukraine needs to complete the demarcation of borders with neighboring countries and get the agreement of neighboring countries, including Russia, on their common border. It is necessary to document everything and sign treaties with all neighboring states. The European Union pledged its support to Ukraine on this important issue and decided to provide full technical assistance. But will Russia sign a border treaty with Ukraine? No of course not Since Russia is the legal successor of the USSR (this is confirmed by the decisions of international courts on property disputes between the former USSR and foreign countries), the lands on which Ukraine, Belarus and Novorossiya are located belong to Russia, and no one has the right to be without Russia's consent to dispose of this area. Basically, now all Russia has to do is declare that this area is Russian and that everything that happens in this area is an internal Russian affair. Any interference will be seen as a measure against Russia. Based on that, they can nullify the May 25, 2014 elections and do whatever the people want! According to the Budapest Memorandum and other agreements, Ukraine has no borders. The state of Ukraine does not exist (and never did!)..'' Alexandre Panin}\\
\textbf{Detected languages:} por: 100.0\% \\
\subsection{Example \#5}
\subsubsection{Fact-check}
\textbf{ID:} 61827 \\
\textbf{Published at:} 2019-11-26 \href{https://periksafakta.afp.com/ini-adalah-foto-seorang-anak-maroko-bukan-mantan-menteri-pendidikan-prancis#7dbc2dfedca6684d5f04acd6287f479f}{periksafakta.afp.com} \\
\\ \textbf{Claim}\\
\textbf{Original text:} \textit{Gadis gembala di Maroko menjadi menteri pendidikan Prancis setelah dewasa}\\
\textbf{Translated text:} \textit{Shepherd girl in Morocco becomes French education minister as an adult}\\
\textbf{Detected languages:} msa: 100.0\% \\
\\ \textbf{Title}\\
\textbf{Original text:} \textit{Ini adalah foto seorang anak Maroko, bukan mantan menteri pendidikan Prancis}\\
\textbf{Translated text:} \textit{This is a photo of a Moroccan child, not a former French education minister}\\
\textbf{Detected languages:} msa: 100.0\% \\

\subsubsection{Social Media Post}
\textbf{ID:} 10815 \\
\textbf{Published at:} Facebook \\
\textbf{Verdicts:} None \\
\\ \textbf{Main text} \\
\textbf{Original text:} \textit{Gadis yg disebelah kiri mengiring domba di maroko, wanita yg disebelah kanan adalah gadis yg sama 20 thn kemudian sbg mentri pendidikan prancis. Jgn pernah berhenti bermimpi dan tdk pernah berhenti bekerja keras utk impian anda.... *** Enerjik. Itulah gambaran sosok Najat Vallaud-Belkacem.  Dulunya, dia memakai baju seadanya dengan rambut dikucir ekor kuda, membawa tongkat, dan menggembalakan domba.  Sehari-hari dia adalah seorang gadis gembala di sebuah desa kecil di dekat Nador, Maroko. Saat itu tidak ada yang menduga bahwa kehidupannya ketika dewasa akan berubah jauh lebih baik. Menjadi menteri pendidikan dan penelitian Prancis. Tentu saja posisi itu tidak begitu saja datang dari langit. Belkacem berusaha ekstrakeras untuk meraihnya. Di kamusnya, tak ada yang tidak bisa diwujudkan. Dulu, ketika dia ingin berkuliah di Paris Institute of Political Studies, guru sekolahnya melarangnya mendaftar. Alasannya, sekolah itu mahal sekaligus susah untuk dimasuki. Namun, langkah anak kedua di antara tujuh bersaudara tersebut tak surut. Belkacem tetap mendaftar, belajar mati-matian, dan akhirnya diterima. Dia juga harus bekerja paro waktu di dua tempat untuk membayar biaya kuliahnya. Di kampus itu pula, dia bertemu dengan Boris Vallaud yang kini menjadi salah seorang penasihat Presiden Prancis Francois Hollande. Mereka sama-sama aktif di Partai Sosialis. Keduanya menikah pada 27 Agustus 2005. Jauh sebelum itu, Belkacem juga sudah terbiasa hidup keras. Saat berusia empat tahun, ayahnya memboyong dia, ibu, dan kakak tertuanya, Fatiha, ke Amiens, kawasan pinggiran Prancis. "Ayah saya tak punya masalah. Tapi, kami, saya, ibu, dan kakak, mati-matian beradaptasi dengan kehidupan baru," katanya seperti dikutip Vogue. Dia bahkan sempat terheran-heran saat melihat mobil. Hal langka di negara asalnya. Belum lagi diskriminasi yang datang dari lingkungan sekitarnya. Bahkan saat dia sudah menjadi anggota parlemen di Rhone-Alpes. Dalam sebuah tulisan, Belkacem bercerita, waktu itu dirinya mengadakan perjamuan makan malam dan mengundang tamu yang belum terlalu mengenalnya. Ketika tamu itu datang, Belkacem menyambut dan membantunya melepaskan mantel. Tamu itu lantas bertanya di mana sang pemilik rumah. "Hingga saat ini di Prancis, kalau ada perempuan dengan kulit berwarna yang membuka pintu rumah di kawasan mewah, selalu dianggap pembantu," tulis ibu si kembar Louis-Adel Vallaud dan Nour-Chloe Vallaud tersebut. Sejak saat itu, dia semakin mantap mengabdikan hidup untuk menghilangkan diskriminasi. Sorotan terhadap karir gemilang Belkacem mulai terjadi saat Presiden Francois Hollande menunjuknya sebagai juru bicara pemerintah dan menteri hak-hak perempuan pada 16 Mei 2012. Beberapa bulan setelah itu, Hollande memberinya tanggung jawab untuk memerangi homofobia. Belkacem menjabat menteri pendidikan dan penelitian pada 25 Agustus 2014, dua hari sebelum ulang tahun kesembilan pernikahannya. Penunjukan itu menjadikan dia sebagai menteri pendidikan termuda yang pernah dipunyai Prancis. Terpilihnya Belkacem seakan menjadi bukti bahwa seorang imigran juga bisa menjadi aset yang berharga bagi negara. Apalagi dia adalah seorang muslim. Tentang Belkacem Saat masih kanak-kanak, momen terbaik dalam hidupnya adalah ketika bibliobus (mobil perpustakaan keliling) menyambangi kawasan tempat tinggalnya. Sebab, dia bisa membaca beragam buku. Memiliki dua kewarganegaraan. Salah satunya Maroko karena dia berasal dari sana. Selain itu, Prancis memberinya status warga negara saat masih kuliah. Ia adalah Anak kedua dari tujuh bersaudara, Najat Belkacem lahir di negara Maroko padan 1977 di Bni Chiker, sebuah desa dekat Nador di wilayah Rif. Pada 1982 ia bergabung kembali dengan ayahnya, seorang pekerja bangunan, dengan ibunya dan kakaknya Fatiha, dan tumbuh di subperkotaan Amiens.[3] Ia lulus dari Institut d'études politiques de Paris (Institut Studi-Studi Politik Paris) pada 2002. Di Institut ia bertemu Boris Vallaud, yang menikah dengannya pada 27 Agustus 2005.[4] Ia masuk Partai Sosialis pada 2002 dan bergabung dengan tim Gérard Collomb, Walikota Lyon, pada 2003 untuk menjalankan demokrasi lokal yang kuat, perlawanan melawan diskriminasi, mempromosikan hak-hak warga sipil, dan akses untuk pekerjaan dan perumahan. Terpilih dalam Dewan Wilayah Rhone-Alpes pada 2004, ia mengetuai Komisi Budaya, mengundurkan diri pada 2008. Pada 2005, ia menjadi penasihat Partai Sosialis. Pada 2005 dan 2006 ia menjadi kolumnis program kebudayaan C'est tout vu di Télé Lyon Municipale bersama dengan Stéphane Cayrol. Pada Februari 2007 ia bergabung dalam tim kampanye Ségolène Royal sebagai jurubicara, bersama dengan Vincent Peillon dan Arnaud Montebourg. Pada Maret 2008 ia ter[ilih menjadi conseillère générale departemen Rhône dalam pemilihan kantonal dengan 58.52\% suara pada putaran kedua, dibawah spanduk Partai Sosialis di kanton Lyon-XIII. Pada 16 Mei 2012, ia dilnatik pada kabinet Presiden Perancis François Hollande sebagai Menteri Hak-Hak Wanita dan jurubicara pemerintahan. https://m.facebook.com/story.php…}\\
\textbf{Translated text:} \textit{The girl on the left is herding sheep in Morocco, the woman on the right is the same girl 20 years later as the French Minister of Education. Never stop dreaming and never stop working hard for your dreams.... *** Energetic. That is the picture of Najat Vallaud-Belkacem. In the past, he wore modest clothes with his hair in a ponytail, carried a stick, and herded sheep. Everyday she is a shepherd girl in a small village near Nador, Morocco. At that time no one expected that his life as an adult would change much for the better. Became the French minister of education and research. Of course that position didn't just come from the sky. Belkacem tried extra hard to reach it. In his dictionary, there is nothing that cannot be realized. In the past, when he wanted to study at the Paris Institute of Political Studies, his school teacher forbade him to enroll. The reason, the school is expensive and difficult to enter. However, the step of the second child among the seven siblings did not subside. Belkacem continued to apply, studied hard, and was finally accepted. He also had to work part-time at two places to pay for his tuition. On the same campus, he met Boris Vallaud, who is now an adviser to French President Francois Hollande. They are both active in the Socialist Party. The two were married on August 27, 2005. Long before that, Belkacem was also used to living hard. When he was four years old, his father took him, his mother and eldest sister, Fatiha, to Amiens, a suburb of France. "My father had no problems. But, we, me, mother and brother, are desperately adapting to a new life," he was quoted as saying by Vogue. He even had time to be surprised when he saw the car. A rare thing in their home country. Not to mention the discrimination that comes from the surrounding environment. Even when he was already a member of parliament in the Rhone-Alpes. In an article, Belkacem recounted that at that time he held a dinner banquet and invited guests who did not know him well. When the guest arrived, Belkacem greeted him and helped him take off his coat. The guest then asked where the owner of the house was. "Until now in France, if a woman of color opened the door to a house in a luxury area, it was always considered a maid," wrote the mother of twins Louis-Adel Vallaud and Nour-Chloe Vallaud. Since then, he has been steadily devoting his life to eliminating discrimination. The spotlight on Belkacem's illustrious career began when President Francois Hollande appointed him as government spokesman and minister for women's rights on 16 May 2012. Months after that, Hollande gave him the responsibility to fight homophobia. Belkacem took office as minister of education and research on August 25, 2014, two days before her ninth wedding anniversary. The appointment makes him the youngest education minister France has ever had. The election of Belkacem seems to be proof that an immigrant can also be a valuable asset for the country. Moreover, he is a Muslim. About Belkacem When he was a child, the best moment in his life was when a bibliobus (mobile library car) visited the area where he lived. Because, he can read a variety of books. Have dual citizenship. One of them is Morocco because he is from there. In addition, France gave him the status of a citizen while still in college. The second of seven children, Najat Belkacem was born in Morocco in 1977 in Bni Chiker, a village near Nador in the Rif region. In 1982 he rejoined his father, a construction worker, with his mother and sister Fatiha, and grew up in the suburb of Amiens.[3] He graduated from the Institut d'études politiques de Paris (Paris Institute of Political Studies) in 2002. At the Institute he met Boris Vallaud, whom he married on 27 August 2005.[4] He joined the Socialist Party in 2002 and joined the team of Gérard Collomb, Mayor of Lyon, in 2003 to promote strong local democracy, fight against discrimination, promote civil rights, and access to jobs and housing. Elected to the Rhone-Alpes County Council in 2004, he chaired the Culture Commission, resigning in 2008. In 2005, he became an adviser to the Socialist Party. In 2005 and 2006 he was columnist for the cultural program C'est tout vu at Télé Lyon Municipale together with Stéphane Cayrol. In February 2007 he joined the Ségolène Royal campaign team as a spokesperson, along with Vincent Peillon and Arnaud Montebourg. In March 2008 he was elected conseillère générale of the Rhne department in cantonal elections with 58.52\% of the vote in the second round, under the banner of the Socialist Party in the canton of Lyon-XIII. On 16 May 2012, she was appointed to the cabinet of French President François Hollande as Minister of Women's Rights and spokesperson for the government. https://m.facebook.com/story.php…}\\
\textbf{Detected languages:} msa: 100.0\% \\

\end{document}